%% file: main.tex
\documentclass[letterpaper, 10 pt, conference]{ieeeconf}  

\IEEEoverridecommandlockouts                              

\usepackage{times}

\usepackage{multicol}
\usepackage[bookmarks=true]{hyperref}
\usepackage{subfiles}
\usepackage{amsmath}
\usepackage{amsfonts}
\usepackage{bm}
\usepackage{booktabs}
\usepackage{comment}
\usepackage{tablefootnote}
\usepackage{multirow}
\usepackage{graphicx}
\usepackage{wrapfig}
\usepackage{mdframed}
\usepackage{adjustbox}
\usepackage[table]{xcolor}
\usepackage{comment}
\usepackage{balance}

\usepackage{capt-of}

\input{macros.tex}

\makeatletter
\long\def\@makecaption#1#2{%
\ifx\@captype\@IEEEtablestring\else\@IEEEfigurecaptionsepspace\fi%
\setbox\@tempboxa\hbox{\footnotesize #1.~~ #2}%
\ifdim \wd\@tempboxa >\hsize%
\setbox\@tempboxa\hbox{\footnotesize #1.~~ }%
\parbox[t]{\hsize}{\footnotesize \noindent\unhbox\@tempboxa#2}%
\else%
\ifcenterfigcaptions \hbox to\hsize{\footnotesize\hfil\box\@tempboxa\hfil}%
\else \hbox to\hsize{\footnotesize\box\@tempboxa\hfil}%
\fi\fi%
\ifx\@captype\@IEEEtablestring\@IEEEtablecaptionsepspace\fi}
\makeatother

\makeatletter
\let\@oldmaketitle\@maketitle
\renewcommand{\@maketitle}{\@oldmaketitle
  \begin{center}
    \includegraphics[width=0.88\textwidth]{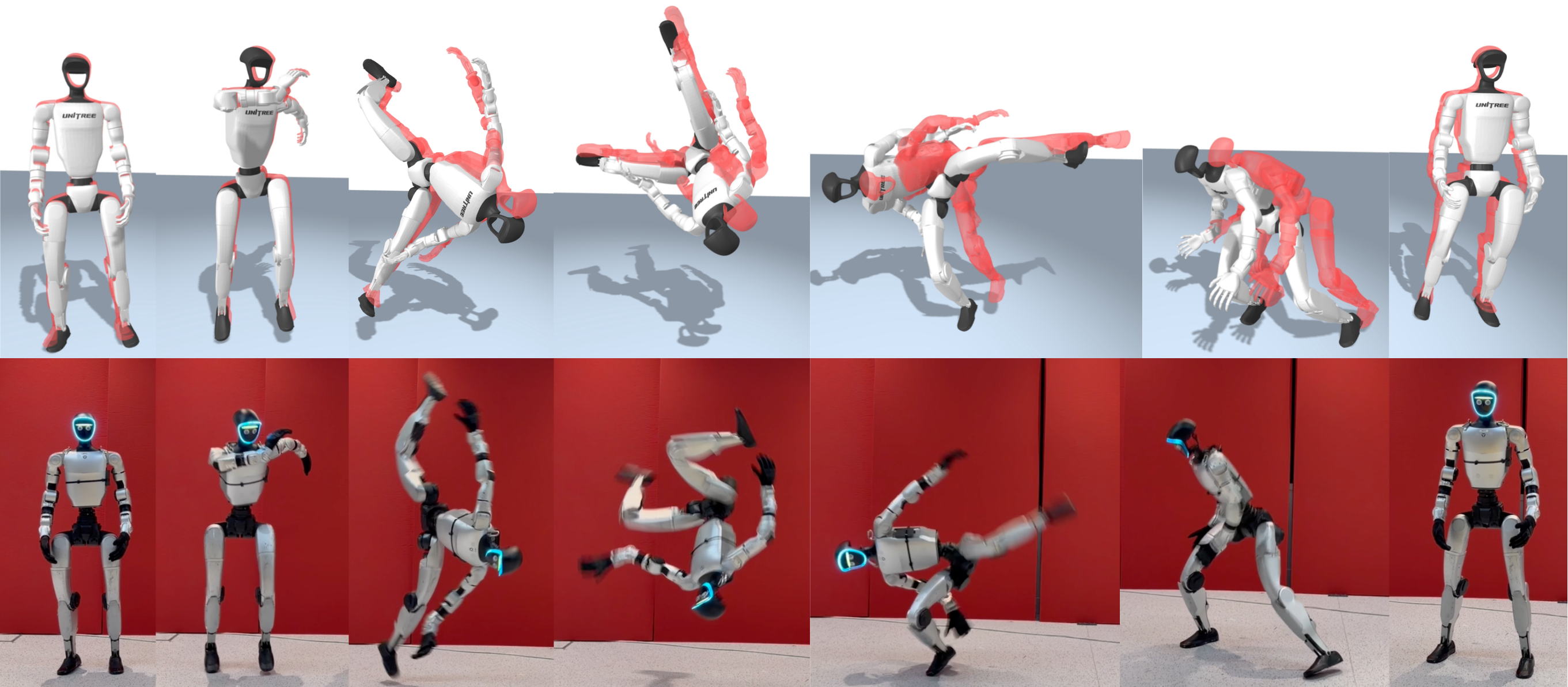}
    \captionof{figure}{\textbf{Dynamically feasible side flip on G1.} \textbf{Top:} overlay of the kinematic reference (red) and dynamically consistent trajectory estimated with our differentiable simulator. \textbf{Bottom:} deployment on the real G1 by a tracking policy trained around the estimated dynamic motion.}
    \label{fig:sideflip}
  \end{center}
}
\makeatother

\begin{document}

\title{\huge Highly-Efficient Differentiable Simulation for Robotics}

\author{Quentin Le Lidec$^{*}$, Louis Montaut$^{*}$, Yann de Mont-Marin$^{*}$, \\ Fabian Schramm, Emilien Bir\'e, and Justin Carpentier
\thanks{$^{*}$Equal contribution. \quad All authors are with Inria, Ecole normale sup\'erieure, PSL Research University, 75005 Paris, France.}%
}

\maketitle

\subfile{abstract}

\IEEEpeerreviewmaketitle

\subfile{sections/1-intro}
\subfile{sections/2-background}
\subfile{sections/3-gradients}

\subfile{sections/4-exp}
\subfile{sections/5-conclusion}

\section*{Acknowledgments}
This work was supported by the European Union's Horizon Europe research and innovation programme through the Marie Sk\l{}odowska-Curie Postdoctoral Fellowship ExTRAORDiNary (grant no.\ 101211945) and the ARTIFACT project (grant no.\ 101165695). Additional support was provided by the French government through the ``PR[AI]RIE-PSAI'' AI Cluster (ANR-23-IACL-0008), managed by the Agence Nationale de la Recherche, and through the France 2030 Organic Robotics Program (PEPR O2R) and the PIQ program, the latter managed by the Agence de Programme du Num\'erique.
Views and opinions expressed are those of the author(s) only and do not necessarily reflect those of the European Union or the European Commission. Neither the European Union nor the European Commission can be held responsible for them.

{
\balance
\bibliographystyle{IEEEtran}
\bibliography{bibliography}
}

\clearpage
\onecolumn
\appendices
\subfile{supp-mat}
\end{document}

%% file: macros.tex
\newcommand{\bmv}{\bm{v}}
\newcommand{\bmq}{\bm{q}}

\newcommand{\RR}{\mathbb{R}}

\definecolor{pastelblue}{rgb}{0.68, 0.78, 0.81}

%% file: abstract.tex
\begin{abstract}
Robotics simulators have improved significantly in computational speed and scalability, enabling them to generate years of simulated data for complex systems in minutes or hours.
Despite these advances, efficiently and accurately computing simulation derivatives remains an open challenge.
Addressing this would accelerate the convergence of reinforcement learning and trajectory optimization algorithms, particularly for contact-rich problems.
This paper introduces a unifying framework for robotic simulation that accounts for all factors, including dynamics, collisions, and friction.
The resulting algorithm computes analytical derivatives of the simulation by implicit differentiation, explicitly handling the intrinsic non-smoothness of the collision and frictional stages while exploiting the sparsity induced by the multi-body structure.
Benchmark results demonstrate state-of-the-art performance, with timings ranging from $5\,\mu$s for a 7-dof manipulator to $95\,\mu$s for a 36-dof humanoid, an improvement of at least two orders of magnitude over alternative methods.
Implemented in C++, the code will be open-sourced after the review process to support applications such as simulation-driven learning and real-time control.

\end{abstract}

%% file: sections/1-intro.tex
\section{Introduction}
\label{sec:intro}

Recent progress in reinforcement learning and trajectory optimization in robotics extensively relies on simulation.
Additional information, such as simulator derivatives, can be leveraged to accelerate the convergence.
However, simulating robotics systems interacting with their environment induces a sequence of nonsmooth operations. 
Typically, collision detection in simulators is intrinsically nonsmooth (e.g., contact points may jump from one vertex to another when the geometry orientation changes slightly), and frictional contact dynamics correspond to nonsmooth problems (e.g., when a cube transitions from sticking to sliding).

Several approaches have been envisaged to estimate simulator derivatives.
Mordatch et al.~\cite{mordatch2012discovery} leverages MuJoCo~\cite{todorov2012mujoco} and finite differences to discover new behaviors.
However, computing gradients via finite differences requires as many forward dynamics calls as the number of parameters to differentiate, which quickly becomes prohibitive.
%
DiffTaichi~\cite{hu2019difftaichi} and NeuralSim~\cite{heiden2021neuralsim} propose exploiting automatic differentiation to differentiate through simplified contact models and geometries.
In this vein, Brax~\cite{freeman2021brax} and MuJoCo MJX build on JAX~\cite{jax2018github} to leverage its automatic differentiation and hardware acceleration for computing gradients through the computational graph. 
Because collision detection and contact forces involve iterative algorithms, gradient computation cost scales with the number of iterations during forward dynamics evaluation.
Alternatively, inspired by differentiable optimization~\cite{blondel2024elements}, multiple works propose to apply implicit differentiation to a linear complementarity problem~(LCP) \cite{de2018end,geilinger2020add,werling2021fast} or mixed linear complementarity problem~(MLCP) \cite{Qiao2021Efficient}, relaxing the original nonlinear complementarity problem~(NCP).
Implicit differentiation has been extended to the NCP case in \cite{howell2022dojo, lelidec2021differentiable}, but it remains inefficient, as it does not, for instance, exploit the structure induced by the kinematic chain. Other approaches, such as \cite{xu2021diffsim}, rely on compliant contact models and focus on differentiating with respect to morphological parameters. For a comprehensive study of different contact models and the impact of relaxing the original NCP on gradients, see \cite{lelidec2023contact,zhong2022differentiable}.\\

\noindent \textbf{Contributions.} 
In this paper, we present a comprehensive framework for differentiable simulation that combines differentiable rigid-body dynamics, differentiable collision detection, and differentiable contact resolution.
We introduce an implicit differentiation scheme to compute the gradients of the NCP associated with the frictional contact problem without relaxation, and then chain it with rigid-body dynamics algorithms to exploit the problem's kinematic sparsity more effectively.
Our approach achieves substantial computational speedups, with gradient computations up to 100 times faster than current state-of-the-art methods in robotics, while avoiding any physical relaxation or geometrical approximation on meshes.
We validate our method on a unified optimal estimation problem through contact, instantiated for estimating initial conditions, inverse dynamics, and recovering a dynamically feasible humanoid side flip from a kinematic reference, which we execute on hardware.
We then show that our gradients improve sample efficiency during training in a policy-learning context.\\

\noindent  \textbf{Paper organization.}
In Sec.~\ref{sec:background}, we provide a background on collision detection, frictional dynamics, rigid multi-body dynamics, and implicit differentiation techniques.
Sec.~\ref{sec:gradients} corresponds to the core contribution of this paper. 
We specify the computational graph of a physics engine and explain how combining gradients of rigid-body dynamics, collision detection, and contact forces makes the simulation end-to-end differentiable.
Additionally, we demonstrate how implicit differentiation and rigid body algorithms can be leveraged to efficiently compute the derivatives of multibody frictional contact problems, including contributions from collision geometry.
In Sec.~\ref{sec:exp}, the efficiency of our approach is benchmarked on several advanced robotics systems.
We also leverage our differentiable physics engine to tackle various estimation problems and to efficiently learn control policies.
Sec.~\ref{sec:limitations} discusses this work's limitations and future developments in model-based approaches for robotics.

%% file: sections/2-background.tex
\section{Background}
\label{sec:background}
This section reviews the three fundamental stages of modern robotic simulators: collision detection, contact modeling, and multibody dynamics.
We also review implicit differentiation, which forms the core of our approach.

\subsection{Collision detection}
\label{subsec:collision-detection}
The collision detection phase identifies the contact points between the colliding geometries composing a simulation scene.
Given two shapes and their relative poses, a collision detection algorithm (e.g., GJK \cite{gilbert1988fast} combined with EPA \cite{van_den_bergen_proximity_2001}) computes a contact point and a contact normal, corresponding to the direction separating the two bodies with minimal displacement.
We define the contact frame $c$ with its origin at the contact point, and the Z axis aligned with the contact normal.
Collision detection algorithms often assume convex geometries; however, existing algorithms \cite{wei2022coacd} can be used during an offline preprocessing phase to decompose nonconvex shapes into convex sub-shapes.

Collision detection is inherently nonsmooth for non-strictly convex geometries \cite{escande2014strictly}.
Specifically, this results in discontinuous contact points and normals.
Thus, differentiating the contact point, normals, and frames w.r.t. the body poses is challenging.
\cite{tracy2023differentiable} uses a smooth approximation of the bodies to calculate the contact frame Jacobians. 
In contrast, \cite{montaut2023differentiable} employs a randomized smoothing approach to compute the derivatives of contact points and normals.

\subsection{Frictional contact dynamics}
\label{subsec:background-frictional-contact}
Given a contact frame between two bodies, let $\bm{\lambda}$ denote the contact force and $\bm{\sigma}$ the contact velocity.
The Signorini condition provides a complementarity constraint \mbox{$0\leq \bm{\lambda}_N \perp \bm{\sigma}_N \geq 0$}, ensuring the normal force is repulsive, bodies do not interpenetrate further, and no simultaneous separation motion and contact force exist.
The maximum dissipation principle (MDP) combined with the frictional Coulomb law \mbox{$\|\bm{\lambda}_T\|\leq \mu \lambda_N$} of friction $\mu$ states that \mbox{$\bm{\lambda}_T \in \operatorname{argmax}_{y, \|y\|\leq \mu \bm{\lambda}_N} -y^\top\bm{\sigma}_T$} maximizes the power dissipated by the contact.
These three principles are equivalent to the following nonlinear complementarity problem (NCP)
\begin{align}
    \label{eq:NCP}
    \mathcal{K}_\mu &\ni \bm{\lambda} \perp \bm{\sigma} + \Gamma_\mu (\bm{\sigma}) \in \mathcal{K}_\mu^*, \\
    \bm \sigma & = G \bm \lambda + g, \nonumber
\end{align}
where $G$ is the so-called Delassus matrix~\cite{delassus1917memoire} that gives the system inverse inertia projected on the contacts. It is a linear operator mapping contact forces to contact velocities. 
$\bm g$ is the free velocity of the contact.
$\mathcal{K}_\mu$ is a second-order cone with aperture angle $\operatorname{atan}(\mu)$, $\mathcal{K}_\mu^* =\mathcal{K}_{1/\mu}$ its dual cone and $\Gamma_\mu(\bm{\sigma})=[0,0,\mu\|\bm{\sigma}_T\|]$ is the so-called De Saxcé correction~\cite{acary2017contact,desaxce1998bipotential} enforcing the Signorini condition~\cite{acary2017contact,lelidec2023contact}.
Problem \eqref{eq:NCP} can be solved by interior point methods~\cite{howell2022dojo}, projected Gauss-Seidel~\cite{jourdan1998gauss} or ADMM-based approaches~\cite{acary2017contact,simplecontacts2024,tasora2021solving}.

\subsection{Multibody frictional contact dynamics}
\label{subsec:background-rigidbody}

We briefly introduce rigid body contact simulation, a core component of physics engines. 
For more detail, see \cite{lelidec2023contact}.
Let $\bm{q} \in \mathcal{Q} \cong \mathbb{R}^{n_q}$ denote the joint position vector, with $\mathcal{Q}$ the configuration space, i.e., the space of minimal coordinates.
The equations of a constrained motion read
\begin{equation}
    \label{eq:constrained_motion}
    M(\bm{q})\bm{\dot{v}} +b(\bm{q},\bm{v}) - \bm{\tau} = J_c^\top(\bm q, c(\bm q)) \bm{\lambda},
\end{equation}
where we denote by $\bm{v} \in \mathcal{T}_q\mathcal{Q} \cong \mathbb{R}^{n_v}$ and $\bm{\tau} \in \mathcal{T}_q^*\mathcal{Q} \cong \mathbb{R}^{n_v}$ the joint velocity
vector and the joint torque vector.
$\dot{\bm{v}}$ is the time derivative of $\bm{v}$.
$M(\bm{q})$ is the joint-space inertia matrix, and $b(\bm{q}, \bm{v})$ includes terms related to the gravity, Coriolis, and centrifugal effects. 
$J_c(\bm q, c(\bm q))$ is the contact Jacobian associated with the contact frame $c(\bm q)$ given by the collision detection on the system bodies using the configuration $\bm q$.
In the following, we drop the parameter dependence when it is explicitly stated.

\begin{table}[t]
    \vspace{.9em}
    \caption{Differentiable physics engines for robotics.}
    \label{tab:contact_solvers}
    \centering
    \begin{adjustbox}{max width=\linewidth}
      \begin{tabular}{r|ccc}
        \toprule
                                         Physics engine  &  Contact Model & $\partial$(Contacts) & $\partial$(Collisions)  \\
        \midrule
        \rowcolor{pastelblue}
        MuJoCo MJX~\cite{todorov2012mujoco} & CCP  & Auto-diff & meshes${}^*$ + primitives \\
        Nimble \cite{de2018end},\cite{werling2021fast}  & LCP  & Implicit & meshes + primitives \\
        \rowcolor{pastelblue}
        Dojo\cite{howell2022dojo} & NCP  & Implicit &  (not considered) \\
       \textbf{Ours}        & NCP  & Implicit &  meshes + primitives  \\
        \bottomrule
      \end{tabular}
    \end{adjustbox}\\
    \vspace{0.5em}
    ${}^*\text{\footnotesize Performances are degraded for meshes over 20 vertices}$
    \vspace{-.5em}
  \end{table}

To deal with rigid-body dynamics and impacts, we use an impulse-based formulation~\cite{mirtich1995impulse} obtained with the Euler symplectic scheme
\begin{equation}
    \label{eq:discrete_motion}
    \bm v^{+} = \bm v + \Delta t \left( \bm {\dot{v}}_f + M^{-1}J_c^\top \bm{\lambda}\right),
\end{equation}
where $\bm {\dot{v}}_f = M^{-1}\left(\bm \tau - b \right)$ is the free acceleration term and $\Delta t$ is the time step. 
The acceleration term $\bm {\dot{v}}_f +M^{-1} J_c^\top \bm{\lambda}$ of \eqref{eq:discrete_motion} correspond to the unconstrained forward dynamics (UFD) with external forces $\bm \lambda$ which can be efficiently computed with the Articulated Body Algorithm (ABA)~\cite{featherstone2014rigid}.
In the remainder of the paper, we use the shorthand $\text{UFD}(\bmq, \bmv, \bm{\tau}, \bm \lambda) = \bm {\dot{v}}_f +M^{-1} J_c^\top \bm{\lambda}$.
Multiplying Eq.~\eqref{eq:discrete_motion} by $J_c$, we recover the contact velocity associated to the contact NCP \eqref{eq:NCP} in the case of multibody dynamics:
\begin{align}
    J_c \bm v^+ = \bm \sigma = G \bm \lambda + g. \label{eq:contact}
\end{align}
This yields the expression of the Delassus matrix \mbox{$G = J_c M^{-1} J_c^\top$} and the free contact velocity vector \mbox{$g = J_c (\bm v + \Delta t \bm {\dot{v}}_f)$}. 
For poly-articulated rigid-body systems, $G$ depends on $\bm q$, and $g$ on $\bm q, \bm v, \bm \tau$.
Thus, the associated NCP is conditioned by $\bm q, \bm v, \bm \tau$. 
Note also that the contact Jacobian $J_c$ depends on $\bm q$, first through the kinematic structure of the system and second, through the contact frame $c(\bm{q})$.

\subsection{Implicit differentiation}
\label{subsec:background-implicit}

As previously mentioned, the physically accurate contact forces denoted by $\bm \lambda^*$ are implicitly defined as the solution of the NCP \eqref{eq:NCP}.
We write $0 = \text{NCP}(\bm \lambda^*; \bm q, \bm v, \bm \tau)$.
By deriving the optimality conditions, the implicit function theorem enables computation of their gradients via implicit differentiation \cite{blondel2024elements}. 
The theorem provides the local derivatives of the solution $\text{d} \bm \lambda^*$ as a linear function of the derivatives of the other variables.

This approach has been applied to differentiate Quadratic Programming (QP) problems in \cite{amos2017optnet} and generalized to convex cone programs \cite{agrawal2019differentiable} and LCPs \cite{de2018end}.
More generally, it enables incorporating optimization layers within the differentiable programming paradigm.
In Sec.~\ref{subsec:NCP-derivatives}, we extend this approach to the NCP case and propose a method to compute the gradients of the contact forces efficiently.

%% file: sections/3-gradients.tex
\section{Efficient differentiable simulation}
\label{sec:gradients}

This section outlines the paper's core contribution: a comprehensive framework for differentiable simulation that integrates differentiable rigid-body dynamics, differentiable collision detection, and differentiable contact resolution. 
We show the link between the derivatives of multibody dynamics, frictional contact dynamics, and collision detection.
We introduce an efficient algorithmic solution to compute the derivatives associated with the contact NCP by solving a reduced system of minimal dimension obtained by implicitly differentiating the contact NCP.

\subsection{Chaining rigid-body dynamics derivatives and NCP derivatives}
\label{sec:grad-end2end}

From Sec.~\ref{sec:background}, the simulation equations~\eqref{eq:discrete_motion} can be restated using unconstrained forward dynamic (UFD) and the solution to the (NCP) problem \eqref{eq:NCP} as
\begin{align}
    \label{eq:NCP_operator}
   0 &= \text{NCP}(\bm \lambda^*; \bm q, \bm v, \bm \tau)\\
    \label{eq:FD}
   \bm v^{+} &= \bm v + \text{UFD}(\bm q, \bm v, \bm \tau, \bm \lambda^*)\Delta t.
\end{align}
Next, we consider forward-mode differentiation setting~\cite{griewank2008evaluating}, i.e., we aim to compute the Jacobian $\frac{\text{d} v^+}{\text{d} \theta}$ where $\theta \in \mathbb{R}^{n_p}$ represents any subset of the inputs $\{ \bm q, \bm v, \bm \tau \}$ or physical parameters.
We can efficiently adapt our approach to reverse mode by applying the computational trick introduced in~\cite{amos2017optnet}.
Differentiating \eqref{eq:FD} leads to
\begin{align}
    \frac{\text{d} \bm v^+}{\text{d} \theta} =&  \frac{\partial \text{UFD}}{\partial \bm \lambda} \frac{\text{d} \bm \lambda^*}{\text{d} \theta}\Delta t\label{eq:FD_grads}\\
    &+\underbrace{
        \frac{\text{d} \bm v}{\text{d}\theta}
        + \left(
            \frac{\partial \text{UFD}}{\partial \bm q} \frac{\text{d} \bm q}{\text{d}\theta} 
            + \frac{\partial \text{UFD}}{\partial \bm v} \frac{\text{d} \bm v}{\text{d}\theta}
            + \frac{\partial \text{UFD}}{\partial \bm \tau} \frac{\text{d} \bm \tau}{\text{d}\theta}
        \right)\Delta t
    }_{\left.\frac{\text{d} \bm v^+}{\text{d} \theta}\right|_{\lambda=\lambda^*}}\notag,
\end{align}
where we identify the term $\left.\frac{\text{d} \bm v^+}{\text{d} \theta}\right|_{\lambda=\lambda^*}$ of derivatives, considering that $\bm \lambda^*$ does not vary. The derivatives $\frac{\partial \text{UFD}}{\partial \bm q, \bm v, \bm \tau}$ and $\frac{\partial \text{UFD}}{\partial \bm \lambda} = M^{-1}(\bm q) J_c^\top(\bm q)$ can be efficiently computed via rigid-body algorithms~\cite{carpentier2018analytical} and are, for instance, available in Pinocchio~\cite{carpentier2019pinocchio}. 
At this stage, it is worth noting that $\frac{\partial \text{UFD}}{\partial \bm q}$ also depends on the geometry of the contact through the contact Jacobian $J_c(\bm q, c(\bm q))$. 
The classical ABA rigid body algorithm gives the derivatives related to the first variable, and we need to add a term related to the variations of $J_c$ induced by the variation of the contact point $c(\bmq)$ (see Section~\ref{subsec:colision-correction}).

Computing the sensitivity of the contact forces $\frac{\text{d} \bm \lambda^*}{\text{d}\theta}$ is also challenging as $\bm \lambda^*$ is obtained implicitly by solving a NCP \eqref{eq:NCP} which depends on $\bm q$, $\bm v$ and $\bm \tau$ through $G$ and $g$.
Notably, the solutions of the NCP are intrinsically nonsmooth, corresponding to the solution of a differential inclusion problem~\cite{acary2017contact}. 
To understand the nonsmoothness of the NCP solutions, consider the case of a contact force either (i) saturating the Coulomb cone or (ii) lying strictly inside the cone.
In case (i), the contact force variations must lie on the tangent plane to the cone at this force value, while in case (ii), no restriction on the contact force variations applies.
In these two cases, the derivatives do not lie on the constraint manifolds.
Next, we detail how implicit differentiation of \eqref{eq:NCP_operator} computes them precisely at a limited computational cost.

\subsection{Implicit differentiation of the NCP}
\label{subsec:NCP-derivatives}

The dynamics induced by the NCP \eqref{eq:NCP} is inherently nonsmooth, as it can switch on three modes.
These modes correspond to the active set of \eqref{eq:NCP} and result in different gradients for the contact dynamics. 
Our approach considers scenarios with multiple contact points and requires identifying the mode for each contact.
For clarity purposes, we present the equations for a single contact point in each of the three modes.\\

\noindent
\textbf{Mode 1 - Breaking contact (brk).} 
This mode corresponds to the case where the contact is separating ($\bm{\sigma}_N>0$), which is induced by the Signorini condition that $\bm{\lambda}^*
= 0$.
This mode can be treated separately from the other two modes, as the contact force is zero and the contact point velocity is not constrained, yielding 
\begin{equation}
    \label{eq:breaking-gradient}
    \frac{\mathrm{d}  \bm{\lambda}^*}{\text{d} \theta} = 0.
\end{equation}

\noindent
\textbf{Mode 2 - Sticking contact (stk).} 
In this mode, the contact point is not moving ($\bm{\sigma} = 0$), which yields the same equations as a bilateral constraint of an attached point \mbox{$G \bm{\lambda}^* + g = 0$}.
Differentiating this constraint gives the following linear equations on $\frac{\mathrm{d}  \bm{\lambda}^*}{\text{d} \theta}$
\begin{equation}
    \label{eq:sticking-gradient}
    G\frac{\mathrm{d}  \bm{\lambda}^*}{\text{d} \theta}   =  - \left(\frac{\text{d} G}{\text{d} \theta} \bm{\lambda}^* + \frac{\mathrm{d}g}{\text{d} \theta}\right).
\end{equation}

\begin{figure}[t!]
    \centering
  \includegraphics[width=0.5\linewidth, trim={0pt 0pt 300pt 0pt}, clip]{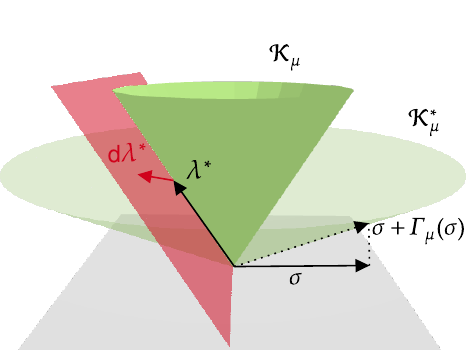} 
  \caption{\textbf{Illustration of the sliding mode.} 
  $\bm \lambda^*$ lives in the boundary of the cone $\mathcal{K}_\mu$ in the direction opposite to $\bm \sigma = \bm \sigma_T$ and the variation $\text{d} \bm \lambda^*$ lies inside the tangent plane.}
  \label{fig:friction_cone}
\end{figure}

\noindent
\textbf{Mode 3 - Sliding contact (sld).} 
In this regime, the contact point moves on the contact surface, which implies a null normal velocity ($\bm{\sigma}_N = 0$) and a non-null tangential velocity ($\|\bm \sigma_T \|>0$).
Moreover, from the MDP, tangential contact forces should lie on the boundary of the cone and in the opposite direction of the tangential velocity (\mbox{$\bm \lambda^*_T = - \mu \bm \lambda^*_N \frac{\bm \sigma_T}{\|\bm \sigma_T \|}$}).
This additionally implies that $ \frac{\mathrm{d}  \bm{\lambda}^*}{\text{d} \theta}$ should be in the plane tangent to the friction cone as illustrated in Fig.~\ref{fig:friction_cone} and allows reducing the search space to a 2D plane via a simple change of variable $ \frac{\mathrm{d}  \bm{\lambda}^*}{\text{d} \theta}  =  R\frac{\mathrm{d}  \bm{\Tilde{\lambda}}}{\text{d} \theta} $ where $R = \begin{pmatrix}
    \frac{\bm \lambda }{\| \bm \lambda  \|} &  \bm e_z \times \frac{\bm \sigma_T }{\| \bm \sigma_T \|}
\end{pmatrix} \in \mathbb{R}^{3\times2}$.
Therefore, differentiating these equations yields the following conditions on the gradients
\begin{equation}
    \label{eq:sliding-gradient}
    \tilde{G}  \frac{\mathrm{d}  \bm{\Tilde{\lambda}}}{\text{d} \theta}   =  - {R}^\top P \left(\frac{\mathrm{d}G }{\text{d} \theta} \bm{\lambda}^* + \frac{\mathrm{d}g }{\text{d} \theta}  \right),
\end{equation}
where:
$P =  \begin{pmatrix}  H(\bm{\sigma}_T ) & 0_{2\times 1} \\
    0_{1\times 2} & 1 
    \end{pmatrix}
    \in \mathbb{R}^{3\times 3}$,
    $H(x) = \frac{1}{\alpha}\left(\text{Id} - \frac{x}{\|x\|} \frac{x}{\| x\|}^\top\right) \in \mathbb{R}^{2\times2}$, $\tilde{G}  = {R}^\top P G  {R } +   Q$ with $Q = \begin{pmatrix}
        0 & 0 \\
        0 & 1
    \end{pmatrix} \in \mathbb{R}^{2\times 2}$ and $\alpha  =\frac{\|\bm{\sigma}_T \|}{\mu  \bm{\lambda}_N }$. 
We refer to \ref{sec:appendix-sliding} for the detailed derivation.

\subsection{Efficient computation: exploiting kinematic sparsity}
\label{sec:gradients-computation}

First, one should identify the active contact modes to obtain the equations for all contact points.
We denote $\mathcal{A}_\text{brk}$, $\mathcal{A}_\text{stk}$, and $\mathcal{A}_\text{sld}$ as the sets of contact indices corresponding to the breaking, sticking, and sliding contacts, respectively.
The dynamics of the different contacts are coupled through the Delassus matrix $G$.
Thus, we construct a matrix $A \in \RR^{(3n_{stk} + 2n_{sld}) \times (3n_{stk} + 2n_{sld})}$ where from $G$ we remove the blocks related to $\mathcal{A}_\text{brk}$ and modify the lines and columns of $G$ related to $\mathcal{A}_\text{sld}$ by following the pattern of $\Tilde{G}$ presented in \eqref{eq:sliding-gradient}.
Once this is done, the reduced linear system on $X$, the stacking of $\text{d} \bm\lambda^*$ and $\text{d} \Tilde{\bm \lambda}$, is obtained by concatenating the corresponding right-hand side of \eqref{eq:sticking-gradient} and \eqref{eq:sliding-gradient}.
We obtain the linear system corresponding to implicit differentiation
\begin{equation}
    \label{eq:ncp-linear-syst}
    A X = - B\left(\frac{\mathrm{d}G }{\text{d} \theta} \bm{\lambda}^* + \frac{\mathrm{d}g }{\text{d} \theta}  \right),
\end{equation}
where $B$ is block diagonal with identity for blocks of $\mathcal{A}_\text{stk}$ and the basis change $R^\top P$ for blocks of $\mathcal{A}_\text{sld}$.
The complete construction is given in the Appendix~\ref{sec:appendix-NCP_syst}.

Computing the right-hand side of \eqref{eq:ncp-linear-syst} requires evaluating the derivative of $G  \bm \lambda^* + g$ with $\bm \lambda^*$ taken constant:
\begin{align}
\frac{\mathrm{d}G }{\text{d} \theta} \bm{\lambda}^* + \frac{\mathrm{d}g }{\text{d} \theta}=\left.\frac{\text{d} G  \bm \lambda^* + g}{\text{d} \theta}\right|_{\lambda=\lambda^*}.
\end{align}
By recalling that $G  \bm \lambda^* + g = J_c \bm v^+$ (see Eq.~\eqref{eq:contact}) is the contact point velocity, this term corresponds to the derivatives w.r.t. $\theta$ of the contact point velocity with $\bm \lambda^*$ taken constant. Calculating the derivative, we obtain:
\begin{align}
    \frac{\text{d}G}{\text{d} \theta} \bm{\lambda}^* + \frac{\mathrm{d}g}{\text{d}\theta}
    = J_c \left.\frac{\text{d} \bm v^+}{\text{d} \theta}\right|_{\lambda=\lambda^*} + \left.\frac{\text{d} J_c \bm v^+}{\text{d} \theta}\right|_{v=v^+}.
\end{align}
The first term is already computed thanks to the ABA derivatives~\eqref{eq:FD_grads}~\cite{carpentier2018analytical}, and the second term $\left.\frac{\text{d} J_c \bm v^+}{\text{d} \theta}\right|_{v=v^+}$, which is the derivative of the contact velocity with $\bm v^+$ assumed to be constant, can be computed at a reduced cost via the partial derivatives of the forward kinematics \cite{carpentier2018analytical} evaluated in $\bm q, \bm v^+$. 
This allows us to avoid the expensive computation of $\frac{\mathrm{d}G }{\text{d} \theta}$ as it is a tensor in general, while only its product with $\bm \lambda$ is required.
At this stage, note that the term $\left.\frac{\text{d} J_c \bm v^+}{\text{d} \theta}\right|_{v=v^+}$ also depends on the geometry of the contact. Therefore, computing gradients w.r.t. the configuration $\bm q$ requires evaluating an additional term for the variations of $J_c$ induced by the variations of contact points on the local geometries \cite{montaut2023differentiable} and presented in the next subsection.

Finally, to obtain $\frac{\text{d} \bm \lambda^*}{\text{d} \theta}$, we solve the linear system \eqref{eq:ncp-linear-syst} using a QR decomposition of $A$ before projecting back the reduced variables $\text{d}\Tilde{\bm \lambda}$ in $\RR^3$.
At this stage, these gradients are computed based on the current active set and thus do not capture information on the contact-mode boundary. 
Still, this is possible by combining our approach with smoothing techniques explored in \cite{suh2022differentiable,pang2023global, zhang2023adaptive}.

\subsection{Collision detection contribution}
\label{subsec:colision-correction}

The collision detection phase depends on the body poses induced by the configuration $\bm q$. Thus, when $\theta$ depends on $\bm q$, one must consider the variation of the contact location given variations of $\bm q$. Recent work on differentiable collision detection \cite{montaut2023differentiable,tracy2023differentiable} enable the computation of the derivative of the normal and contact point with respect to the poses of the bodies.

By choosing a function that constructs a contact frame $c$ from a contact point and its normal, we compute the derivative of this frame w.r.t. the body poses.
Chaining this derivative with the usual kinematics Jacobian, which relates the variation of $\bm q$ to the variation of body poses, one can obtain $\frac{\text{d} c}{\text{d} \theta}$.
The frame $c$ intervenes in $J_c$ through a change of frame (the adjoint of the placement). 
By leveraging classic spatial algebra formula~\cite{featherstone2014rigid} and associated derivatives~\cite{carpentier2018analytical}, 
one can compute $\frac{\partial J_c^\top \bm \lambda^*}{\partial c}$ and $\frac{\partial J_c \bm v^+}{\partial c}$ (see Appendix~\ref{sec:appendix-coldetection_contrib} for details).
We add the first collision term $\frac{\partial J_c^\top \bm \lambda^*}{\partial c}\frac{\text{d} c}{\text{d} \theta}$ in $\left.\frac{\text{d} \bm v^+}{\text{d} \theta}\right|_{\lambda=\lambda^*}$ and the second collision term $\frac{\partial J_c \bm v^+}{\partial c}\frac{\text{d} c}{\text{d} \theta}$ in $\left.\frac{\text{d} J_c \bm v^+}{\text{d} \theta}\right|_{v=v^+}$ to account for the variation of $J_c$ due to the variation of the contact points. 

The complete details of the terms and simulation derivative $\frac{\text{d} \bm v^+}{\text{d} \theta}$ are reported in Appendix~\ref{sec:appendix-gradient_expr}.

%% file: sections/4-exp.tex
\section{Experiments}
\label{sec:exp}
\begin{figure}
    \centering
    \includegraphics[width=.32\linewidth]{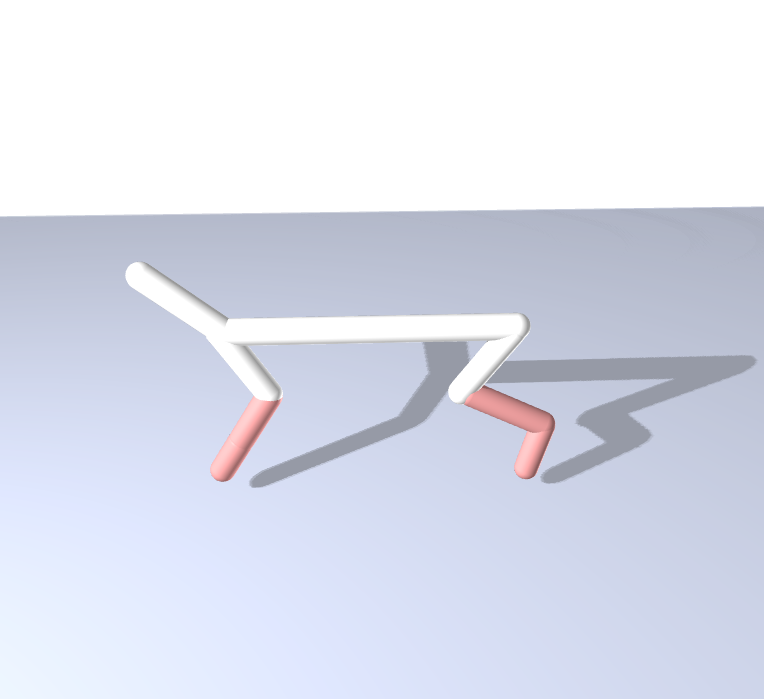}\,
    \includegraphics[width=.32\linewidth]{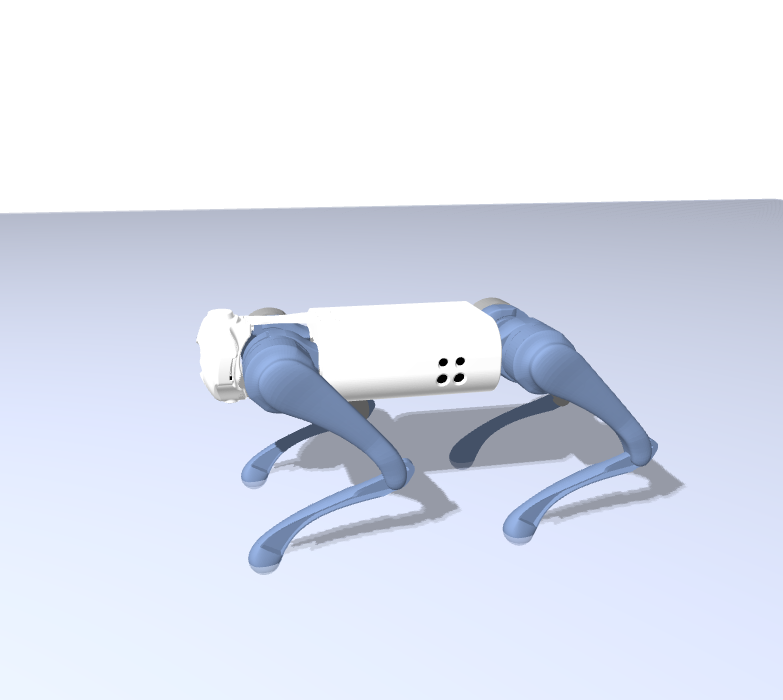}\,
    \includegraphics[width=.32\linewidth]{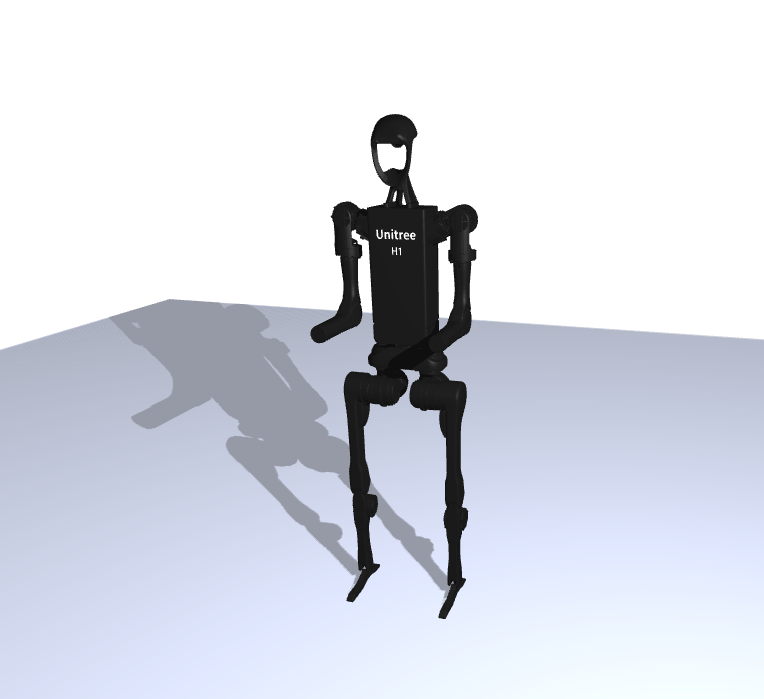}
\caption{\textbf{Evaluated robotic systems.} From geometric primitives (\textbf{Left}, MuJoCo's half-cheetah) to high-dof robots with complex geometries (\textbf{Center}, Unitree Go1; \textbf{Right}, H1).}
    \label{fig:robots}
\end{figure}

We first report computational timings for simulation gradients across a range of robotic systems (Fig.~\ref{fig:robots}), including full-mesh geometries and high-dof humanoids.
Second, we use these gradients to address optimal estimation through contact. We formulate a nonlinear least-squares problem and instantiate it three times at increasing complexity: recovering the initial conditions of a thrown cube,
recovering the actuation torques of a standing quadruped, and estimating a dynamically feasible side flip for a humanoid from a kinematic reference. The latter is executed on hardware.
Finally, we apply our approach to first-order policy learning algorithms to train control policies efficiently. Additional results are available in the companion video.

\noindent \textbf{Implementation details. }
We have implemented our analytical derivatives in C++ for efficiency. 
We leverage open-source software from the community: Eigen~\cite{eigenweb} for efficient linear algebra,  Pinocchio~\cite{carpentier2019pinocchio} for fast rigid body dynamics and their derivatives, and Coal~\cite{hppfcl} for high-speed collision detection. 
We will release the code associated with this paper as open source upon acceptance.
All the experiments are performed on a single core of an Apple M3 CPU.

\subsection{Timing benchmarks}
\label{sec:exp-timings}

\begin{table}[b]
    \vspace{-.65em}
    \caption{\textbf{Details of scenarios for computational timings.} N/A indicates a scenario with no mesh.}
    \label{tab:timings-details}
    \centering
    \begin{adjustbox}{max width=\linewidth}
      \begin{tabular}{>{\raggedleft\arraybackslash}p{1.5cm}||c|c|c|c|c|c}
        \toprule
        & \textbf{Half-cheetah} & \textbf{Humanoid} & \textbf{UR5} & \textbf{Go1} & \textbf{H1} & \textbf{Atlas} \\
        \midrule
        \rowcolor{pastelblue}
        \# DoFs & 12 & 27 & 7 & 18 & 25 & 36 \\
        \# geometries & 9 & 20 & 9 & 39 & 25 & 89 \\
        \rowcolor{pastelblue}
        \# collision pairs & 27 & 161 & 26 & 494 & 255 & 3399 \\
        \# vertices per mesh & N/A & N/A & 200 & N/A & 700 & N/A \\
        \bottomrule
      \end{tabular}
    \end{adjustbox}
\end{table}
\newcommand\ExtraSep
{\dimexpr\cmidrulewidth+\aboverulesep+\belowrulesep\relax}
\begin{table*}[t]
    \centering
    \caption{\textbf{Timing comparison of ours, MuJoCo, and MJX}. Mean and standard deviation in microseconds for simulation, gradient, and finite-difference computation for one simulation step. N/A denotes geometries unsupported by MJX.}
    \begin{adjustbox}{max width=\textwidth}
      \begin{tabular}{r||c|c|c|c|c|c|c}
        \toprule
                                     & \textbf{Half-cheetah} & \textbf{Humanoid}  & \textbf{UR5}  & \textbf{Go1} & \textbf{H1} & \textbf{Atlas} & \textbf{Framework} \\
        \midrule 
        \rowcolor{pastelblue}
        Simulation & $15.8 \pm 7.1$ & $42.6 \pm 24.7$ & $12.3 \pm 6.1$ & $62.6 \pm 18.5$ & $139.3 \pm 125.4$ & $127.3 \pm 32.9$ & \cellcolor{white} \textbf{Ours} \\
        Implicit gradients & $14.7 \pm 7.0$ & $47.9 \pm 23.8$ & $4.1 \pm 3.8$ & $93.0 \pm 32.0$ & $54.3 \pm 34.5$ & $95.2 \pm 37.6$ & \cellcolor{white} Apple M3 CPU \\
        \rowcolor{pastelblue}
        Finite differences & $1.1\mathrm{e}{3} \pm 0.5\mathrm{e}{3}$ & $5.5\mathrm{e}{3} \pm 3.5\mathrm{e}{3}$ & $0.4\mathrm{e}{3} \pm 0.2\mathrm{e}{3}$ & $6.6\mathrm{e}{3} \pm 1.8\mathrm{e}{3}$ & $17.6\mathrm{e}{3}\pm 14.6\mathrm{e}{3}$ & $26.7\mathrm{e}{3} \pm 6.\mathrm{e}{3}$ & \cellcolor{white}\\
        \midrule \midrule
        Simulation & 
        $5.5 \pm 2.0$ & 
        $40.3 \pm 40.1$ & 
        $12.3 \pm 4.0$ & 
        $15.8 \pm 7.0$ & 
        $59.3 \pm 31.0$ & 
        $85.1 \pm 34.4$ & 
        \cellcolor{white} \textbf{MuJoCo} \\
        \rowcolor{pastelblue}
        Finite differences & 
        $0.34\mathrm{e}{3} \pm 0.13\mathrm{e}{3}$ & 
        $2.9\mathrm{e}{3} \pm 0.8\mathrm{e}{3}$ & 
        $0.39\mathrm{e}{3} \pm 0.10\mathrm{e}{3}$ & 
        $9.9\mathrm{e}{3} \pm 0.16\mathrm{e}{3}$ & 
        $5.9\mathrm{e}{3}\pm 1.4\mathrm{e}{3}$ & 
        $54.3\mathrm{e}{3} \pm 1.4\mathrm{e}{3}$ & 
        \cellcolor{white} Apple M3 CPU \\
        \midrule \midrule
        Simulation & $1.0 \pm 0.0$ & $2.3 \pm 0.0$ & N/A & N/A & N/A & N/A & \cellcolor{white} \textbf{MJX} \\
        \rowcolor{pastelblue}
        Autodiff gradients & $3.7 \pm 0.0$ & $103.2 \pm 0.3$ & N/A & N/A & N/A & N/A & \cellcolor{white} Nvidia A100 GPU \\
        \bottomrule
      \end{tabular}
    \end{adjustbox}
    \vspace{-.9em}
    \label{tab:timings}
\end{table*}

The computational efficiency of our approach is evaluated by measuring the average time required to compute the full Jacobian of the simulator, i.e., $\frac{\text{d} \bm v^+}{\text{d} \bm q, \bm v, \bm \tau}$, along a trajectory.
We consider scenarios ranging from simple systems composed of basic geometry primitives (MuJoCo's half-cheetah and humanoid) to more complex, realistic robots with multiple DoFs and complex geometries (UR5, Unitree Go1, and Boston Dynamics Atlas).
Tab.~\ref{tab:timings-details} reports the numbers for the different robots considered.
To stabilize simulation behavior, we compute contact collision patches (each with 4 contact points), substantially increasing the problem size.

\begin{figure}[!t]
    \centering
    \vspace{1em}
    \includegraphics[width=0.9\linewidth]{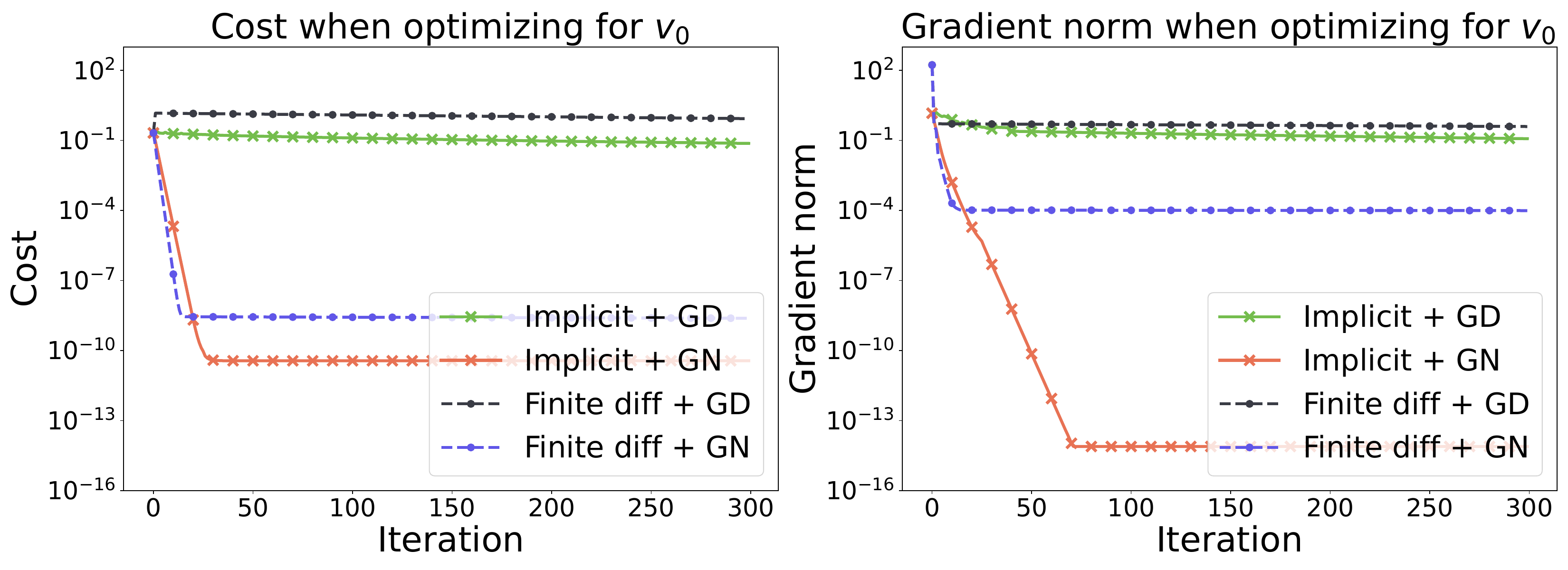}
    \includegraphics[width=0.9\linewidth]{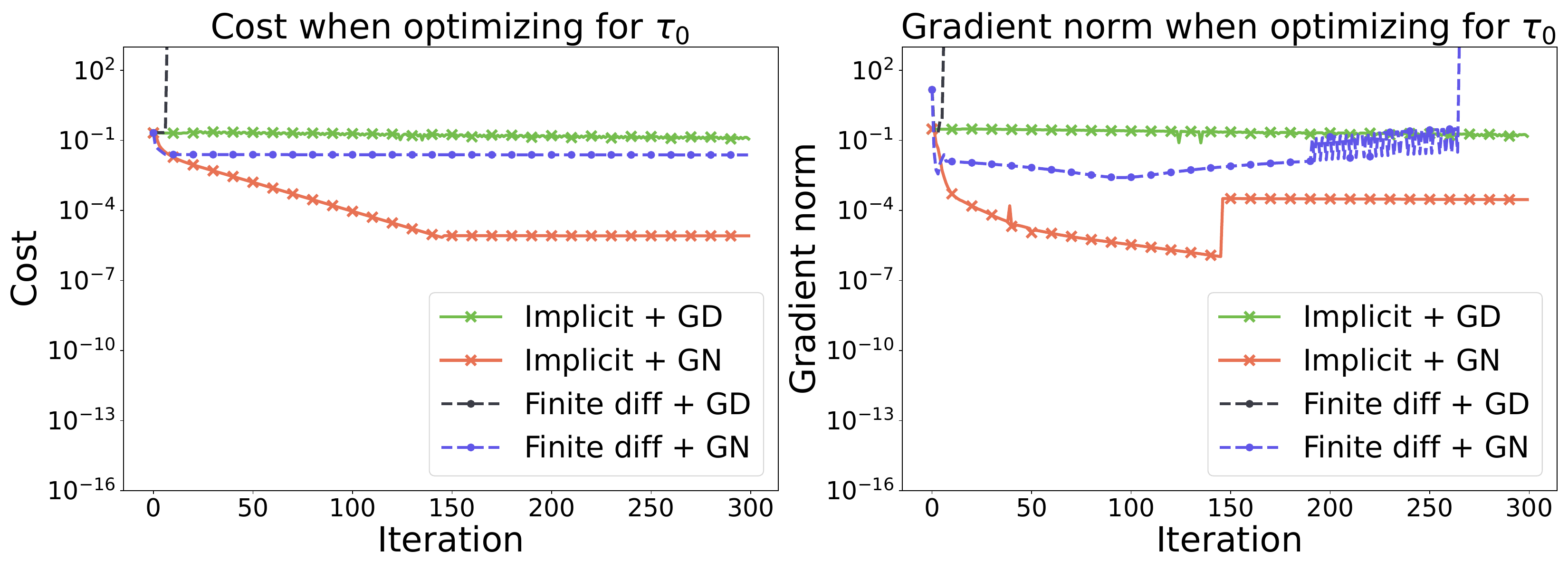}
    \vspace{-.8em}
    \caption{\textbf{Estimation of initial conditions.} Gauss-Newton (GN) algorithm uses efficient implicit differentiation to retrieve the initial velocity $\bm v_0$ and impulse $\bm \tau_0$. In the third and fourth figures, Gradient Descent with finite differences diverges because the estimated gradients are excessively large. When at the boundary of a contact mode, the norm of the finite differences gradients becomes inversely proportional to the step size.}
    \label{fig:traj_id}
    \vspace{-.65em}
\end{figure}

Tab.~\ref{tab:timings} demonstrates computational timings for gradient computation that are of the same order of magnitude as simulation and significantly faster than central finite differences.
As another point of comparison, Nimble~\cite{werling2021fast} requires 1ms and 16ms on half-cheetah and Atlas, corresponding to an approximate speedup factor of 100 for our method.
In Tab.~\ref{tab:timings}, we also compare our approach to MuJoCo MJX GPU simulation pipeline.
The number of gradients computed on the GPU corresponds to the samples generated in one second using all threads of an Nvidia A100.
We find our approach competitive even though it runs on a single CPU core.
Importantly, we achieve this performance gain despite working on the unrelaxed NCP and with full mesh descriptions (cf. Table~\ref{tab:contact_solvers}).
In general, current GPU simulation approaches do not achieve this, as hardware constraints require relaxed physical principles and simplified geometries.

\subsection{Optimal estimation through contact}
\label{sec:exp-sysid}

Denoting by $\bm x_t = (\bm q_t, \bm v_t)$ the state and $\bm f$ the simulation step, the three problems considered here read
\begin{equation}
    \label{eq:estimation}
    \begin{aligned}
        \min_{\bm \theta} \quad & \frac{1}{2} \sum_{t=1}^{T}
        \left\| \bm h(\bm x_t) - \bm y_t \right\|_{\bm W}^2
        + \frac{\rho}{2} \left\| \bm \theta \right\|^2_D \\
        \text{with} \quad &
        \bm x_{0} = \bm g(\bm \theta)\\
        &\bm x_{t+1} = \bm f(\bm x_t, \bm \tau_t(\bm \theta))
    \end{aligned}
\end{equation}
where $\bm y_{1:T}$ are observations, $\bm h$ an observation model, $\bm g$ gives a task-dependent initial condition, $\rho$ a regularization gain, and $\bm \theta$ the estimated quantity, on which the rollout depends through its initial state $\bm x_0$ or its controls $\bm \tau_{0:T-1}$.
The dynamics hold by construction, so \eqref{eq:estimation} is an unconstrained least-squares problem in $\bm \theta$.
This is the classical optimal estimation setting~\cite{rawlings2017model}, solved by Gauss-Newton with a Jacobian obtained by chaining the simulator derivatives along the horizon.
Since contact renders \eqref{eq:estimation} nonsmooth, the accuracy of the estimate depends on the accuracy of these Jacobians, which we verify by comparing the same solver with finite differences in each instance.

\vspace{0.2cm} \noindent  \textbf{Estimating initial conditions.}
We first consider a cube in pose $\bm q_0$ thrown onto the floor, evolving in a sliding mode, whose initial condition 
-- either an initial velocity $\bm \theta = \bm v_0$ or an initial impulse $\bm \theta = \bm \tau_0$ --
has to be estimated from a single terminal pose observation $\bm h (\bm x_T) = \bm q_T$ matching a target observation $\bm y_T = \bm q^*$ after $T$ steps.
Fig.~\ref{fig:traj_id} shows that our implicit gradients reach a solution substantially more precise than the one obtained with finite differences, and that exploiting the full Jacobian in Gauss-Newton reduces the iteration count compared to that of vanilla gradient descent.

\begin{figure}[t]
    \centering
    \includegraphics[width=.9\linewidth]{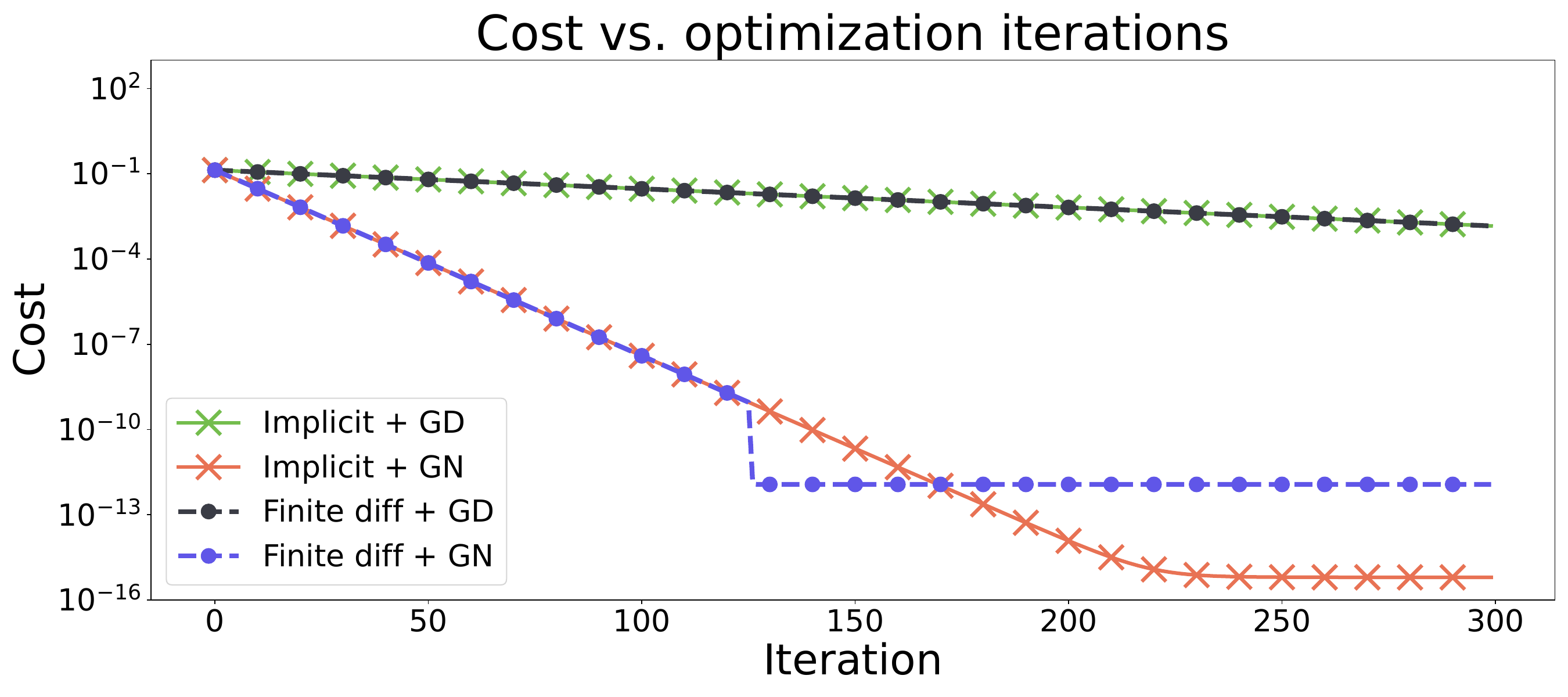}\\
    \includegraphics[width=.9\linewidth]{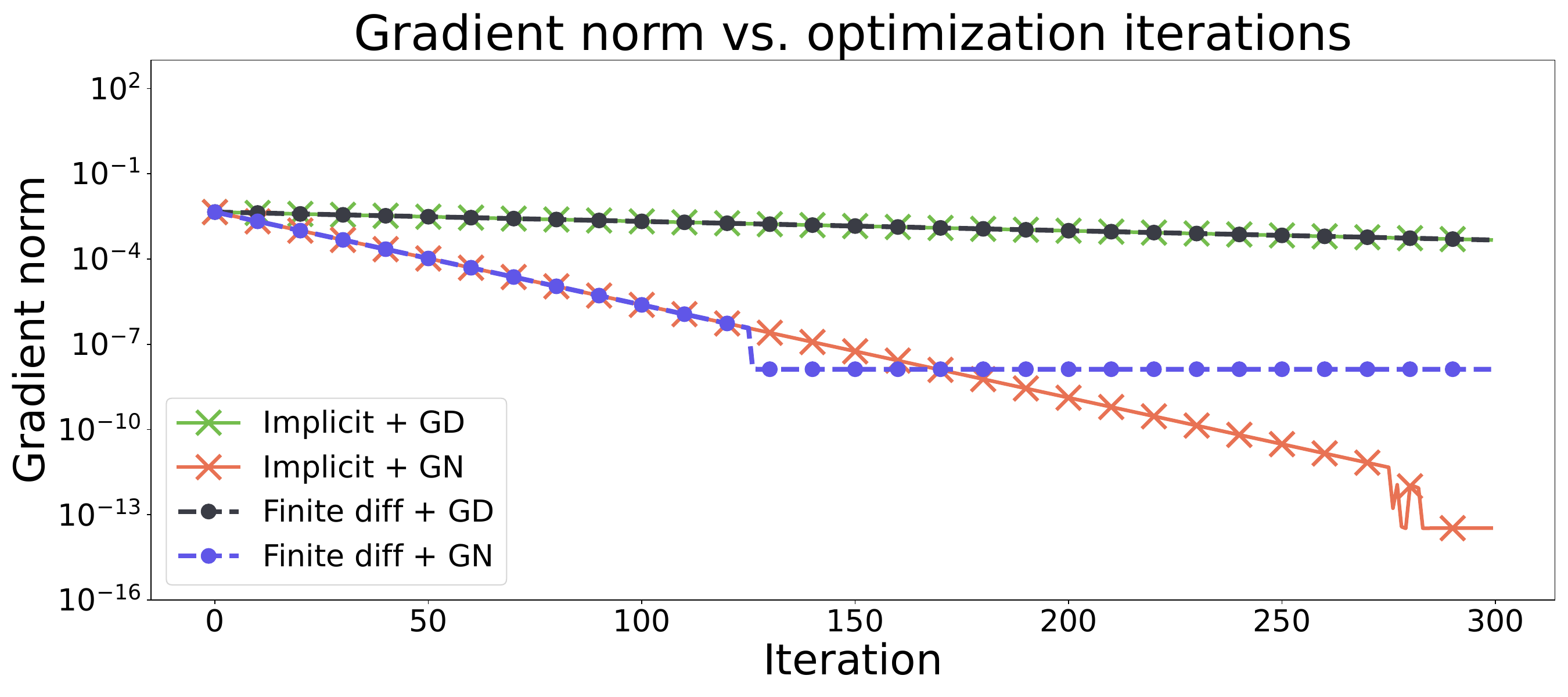}
    \caption{
   \textbf{Contact inverse dynamics.} Cost and gradient norm when recovering the torques that hold an underactuated Go1 quadruped at a null acceleration. Gauss-Newton with our implicit derivatives converges in about ten iterations, to a precision unattainable with finite differences.
    }
    \label{fig:invdyn}
    \vspace{-.75em}
\end{figure}

\vspace{0.2cm} \noindent \textbf{Inverse dynamics through contacts.}
The single-step case $T=1$ of \eqref{eq:estimation} corresponds to inverse dynamics through contacts.
We consider a Unitree Go1 quadruped in a standing position $(\bm q_0, \bm v_0)$ and estimate the actuation torques $\bm \theta = \bm \tau_0$ (mapped to the joints by an actuation matrix $S$)
that realize an observed next velocity $\bm h(\bm x_1) = \bm v_1$ matching a target observation $\bm y_1 = \bm v^*$.
In our experiment, we take $\bm v_0$ and $\bm v^*$ to be null, to estimate torques that hold the underactuated robot in a static pose $\bm q_0$.
As shown in Fig.~\ref{fig:invdyn}, roughly ten Gauss-Newton iterations suffice to reach an error of $1\mathrm{e}{-5}$, again with a precision unattainable using finite differences. Other inverse dynamics experiments are available in Appendix~\ref{sec:appendix-additional-exp}.

\vspace{0.2cm} \noindent \textbf{Dynamically feasible side flip.}
The full-horizon case of \eqref{eq:estimation} consists of recovering an executable motion from a kinematic reference.
We take as input a reference $\bm q^\text{kin}_{1:T}$ of a human side flip retargeted onto the joints of a G1 humanoid.
Such a reference is purely geometric and not directly executable: it ignores the robot's inertia, actuation limits, and the unilaterality of contact forces.
Taking it as the target observation $\bm y_t = \bm q^\text{kin}_t$ for the configuration observation $\bm h(\bm x_t) = \bm q_t$, we estimate the dynamically feasible trajectory -- through torques $\bm \theta = \bm \tau_{0:T-1}$ -- that best explains it. The objective is
\begin{equation}
    \label{eq:retargeting}
    \min_{\bm \tau_{0:T-1}} \frac{1}{2} \sum_{t=1}^{T}
    \left\| \bm q_t \ominus \bm q^\text{kin}_{t} \right\|_{\bm W}^2
    + \frac{\rho}{2} \left\| \bm \tau \right\|_2^2 ,
\end{equation}
where $\ominus$ denotes the difference on the configuration manifold, accounting for the free-flyer joint, and $\bm W$ weights the base pose against the joint angles.
Fig.~\ref{fig:sideflip} (top) shows the estimated trajectory against the reference it explains.
In practice, problem \eqref{eq:retargeting} is solved multiple times over a receding window of size $T = 30$: the first $4$ controls are committed, the window is shifted, and the previous solution warm-starts the next one, following a moving-horizon estimation scheme.
We carry out this procedure using both implicit differentiation and finite differences to build the Jacobians. 
Comparisons are available in Tab.~\ref{tab:gn_iters_walltime}.


\begin{table}[b]
    \vspace{-.5em}
    \caption{\textbf{Optimizer cost: finite-difference (FD) vs.\ implicit derivatives}. Iteration counts are averaged per window, the real-time factor is the wall time needed to estimate one second of motion. FD are $29\times$ slower.}
    \label{tab:gn_iters_walltime}
    \centering
    \begin{adjustbox}{max width=\linewidth}
      \begin{tabular}{r|cc|c}
        \toprule
        & \multicolumn{2}{c|}{\# GN iterations / window} & Real-time factor
        \\
        \cmidrule(lr){2-3}
        & Median & IQR & \\
        \midrule
        \rowcolor{pastelblue}
        \textbf{Implicit (ours)} & 5 & $[3, 6]$ & $7.8 \pm 6.1$ \\
        Finite differences & 5 & $[4, 7]$ & $227 \pm 172$ \\
        \bottomrule
      \end{tabular}
    \end{adjustbox}
\end{table}

\vspace{0.2cm} \noindent \textbf{Deployment on real hardware.}
The estimated trajectory is dynamically feasible but open-loop, and therefore cannot reject model mismatch and disturbances encountered on a real robot. Following standard practice, we obtain a closed-loop controller by training a tracking policy around it with PPO~\cite{schulman2017proximalpolicyoptimizationalgorithms} under domain randomization.
This makes the estimated motion executable on hardware, as shown in Fig.~\ref{fig:sideflip} and the companion video, where the executed motion remains close to the estimate in \eqref{eq:retargeting}.

\subsection{Applications to policy learning}

Model-free reinforcement learning relies on zeroth-order gradient estimation via the policy gradient theorem, often leading to high variance and slow convergence. In contrast, first-order gradient-based optimization using analytical gradients from a differentiable simulator enables more efficient policy updates~\cite{xu2021accelerated, georgiev2024adaptive}.
We integrate our differentiable simulator with the first-order, on-policy algorithm SHAC and compare it to the zeroth-order, on-policy algorithm PPO on the cartpole swing-up and the MuJoCo hopper task. The performances are reported in Fig.~\ref {fig:shac_vs_ppo}. 
Although these systems are relatively small, they are not trivial because of their non-smooth dynamics. 
The cartpole involves joint limits, and the hopper combines joint constraints with ground contact interactions.
Our simulation approach models non-compliant contacts and joint limits without any relaxations~\cite{simplecontacts2024}.
In turn, the absence of relaxation directly affects the smoothness and continuity of the gradients, often making it challenging to leverage them for gradient-based optimization.

The experiments (see Appendix~\ref{sec:appendix-policy} for details)show that using our differentiable simulator with SHAC is more sample-efficient than PPO. However, the training process is noticeably noisier. 
This may stem from the stiffness of our simulator, which models non-smooth constraints directly rather than using commonly employed relaxations or spring-damper systems. 
Although differentiable simulation reduces gradient variance, the resulting gradients can become highly unstable in stiff scenarios. 
Mitigating this instability is challenging \cite{georgiev2024adaptive} and may require effective gradient-smoothing strategies. 
Potential approaches include adaptively scheduling a compliance term for the constraints or leveraging solver methods that inherently apply smoothing, such as interior point solvers \cite{florian2000ipm,howell2022dojo}, to improve stability without sacrificing efficiency.

%% file: sections/5-conclusion.tex
\section{Discussions} 
\label{sec:limitations}
\begin{figure}[t]
    \centering
    \includegraphics[width=.48\linewidth]{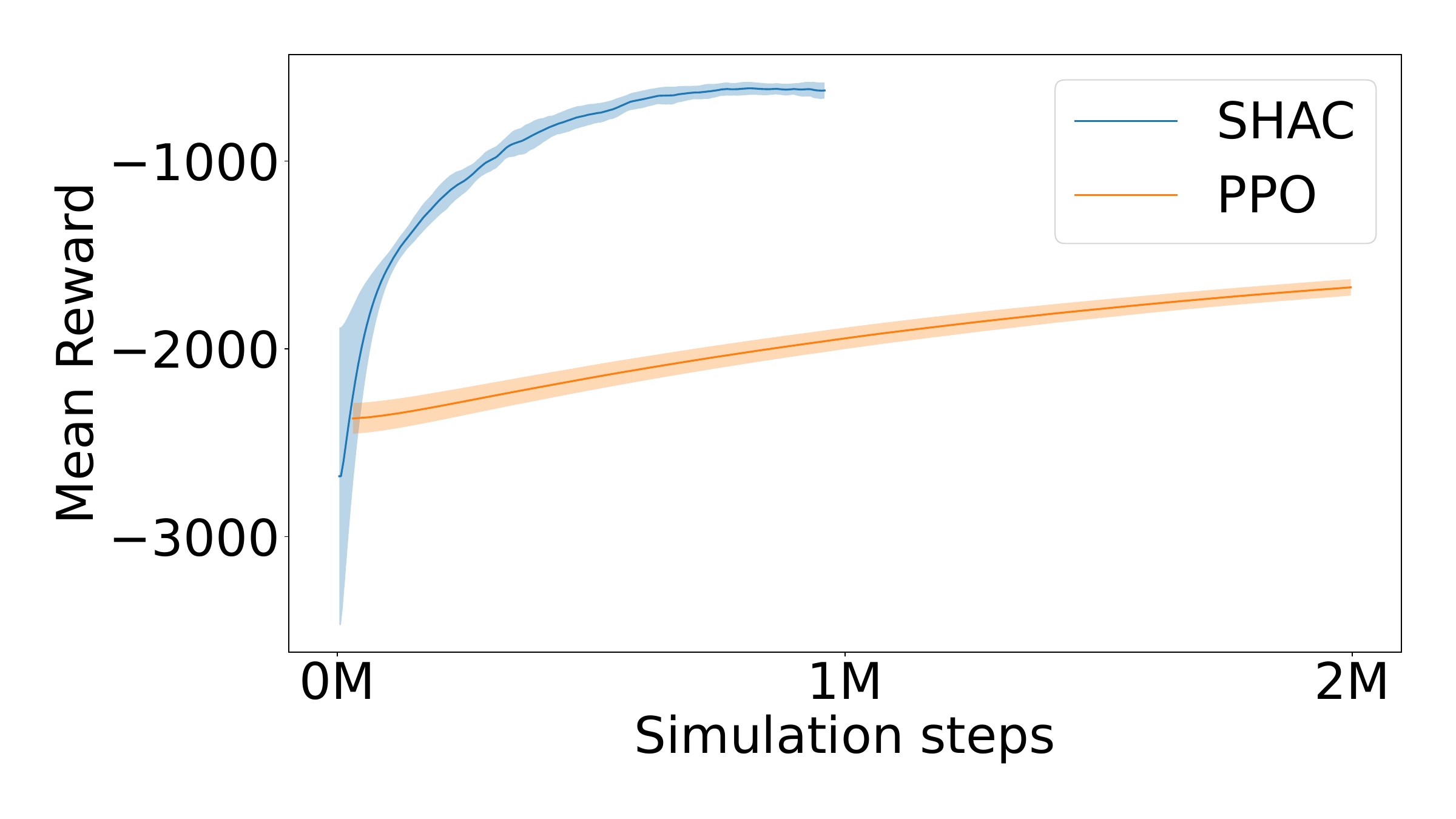}\,
    \includegraphics[width=.48\linewidth]{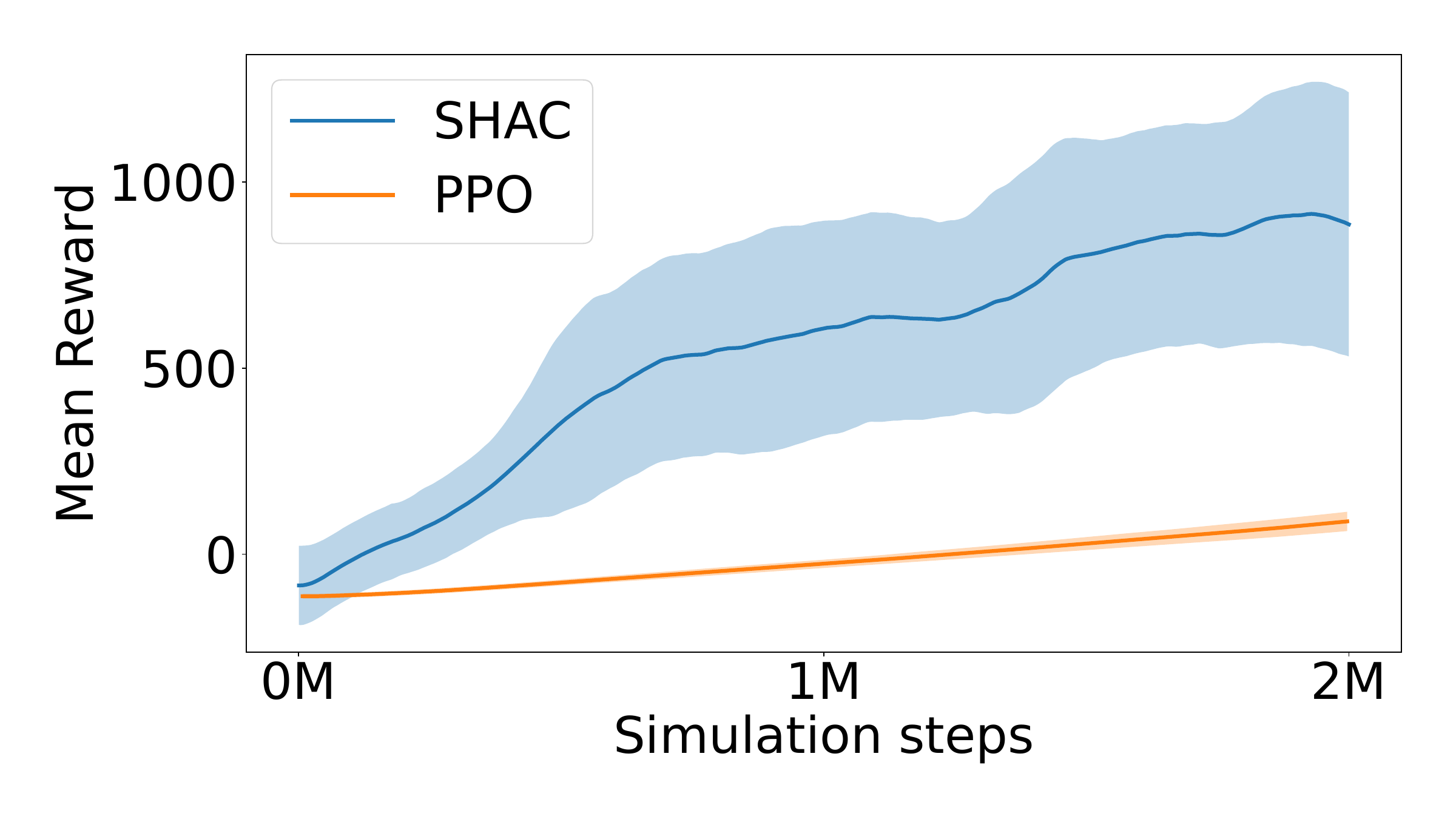}
    \caption{
    \textbf{SHAC vs. PPO} on cartpole swing-up (left) and hopper (right). The SHAC algorithm uses differentiable simulation to improve sample efficiency.
    }
    \label{fig:shac_vs_ppo}
    \vspace{-.75em}
\end{figure}
This paper introduces an end-to-end differentiable physics pipeline for robotics based on the implicit differentiation of the non-relaxed NCP for contacts.
By avoiding relaxation, we prevent nonphysical simulation artifacts.
Moreover, exploiting the sparsity induced by the robot kinematic chains and leveraging derivatives of rigid body algorithms enable state-of-the-art timings, with a speed-up of at least $100$ compared to alternative state-of-the-art solutions.
In an MPC context where the dynamics and their derivatives are evaluated at high frequency, gains in physical realism and efficiency can significantly affect the controller's overall performance.

Yet, because the NCP induces inherently non-smooth dynamics, exploiting its gradients requires dedicated algorithms for downstream optimization tasks.
Our first-order reinforcement learning experiments highlight this challenge: directly using the simulator's raw analytical gradients with NCP-modeled constraints yields efficient but sometimes less stable convergence.
Some previous works~\cite{suh2022bundled, lelidec2024leveraging, pang2023global} have leveraged randomized smoothing techniques to provide smooth gradient estimates from simulation and gradient samples. 
Alternative solutions relax the physics, either explicitly \cite{todorov2012mujoco,mordatch2012discovery,kim2023contact} or implicitly, by leveraging interior-point (IP) methods \cite{howell2022dojo}.

Like existing robotic simulators (e.g., MuJoCo, Bullet, DART), this paper models contact interactions as vanilla 3D contact points, though richer but more complex contact models exist. One promising direction is to extend this work to deformable contact interactions to improve simulator realism, as in~\cite{elandt2019pressure}.

\section{Conclusion}
In future work, we plan to combine our qualitative gradient approach with smoothing techniques to integrate it into recent control frameworks and tackle more complex robotics tasks. 
Specifically, this paper introduces an efficient method for computing physics gradients, which are used in first-order policy learning algorithms such as SHAC~\cite{xu2021accelerated} and AHAC~\cite{georgiev2024adaptive}. The next step is extending our SHAC experiments to more complex real-world robotics systems.
Our approach is not limited to rigid body robots; it can also be leveraged for soft dynamics in general to design and control soft robots~\cite{della2023model} and can be adapted to use an implicit integrator, as in~\cite{castro2023compliant}.
The proposed differentiable simulation of an unrelaxed physics model is a crucial step toward reducing the Sim2Real gap~\cite{hofer2021sim2real}, and future work may adapt learning algorithms to achieve this goal effectively.
Finally, we hope this work will catalyze the robotics and learning communities, motivating the development of new reinforcement learning and trajectory optimization methods that leverage simulation gradients to accelerate the discovery of complex robot movements in contact.

%% file: supp-mat.tex
\section{Sliding mode}
\label{sec:appendix-sliding}
In this section, we detail the equations (Eq.~\ref{eq:sliding-gradient} of the paper) of the sensitivity analysis of the contact force in the case of a sliding mode ($\|\bm \sigma_T\| > 0$).
In sliding mode, both the contact forces and the contact point velocity are on the border of their cone. Thus, from the NCP we have:
  \begin{subequations}
  \begin{alignat}{2}
     \|\bm{\lambda}_T\| = \mu \bm{\lambda}_N \\
    \bm\sigma_N=0\\
      {\bm{\lambda}} ^ \top \big ( \bm \sigma  + \Gamma_\mu ( \bm \sigma )\big )  = 0\\
      \bm \sigma = G \bm \lambda + g,
  \end{alignat}
  \end{subequations}
  which is equivalent to:
  \begin{subequations}
  \begin{alignat}{2}
     \bm{\lambda}_T =  - \mu \bm{\lambda}_N \frac{\bm \sigma_T}{\|\bm \sigma_T\|}\\
       \bm \sigma_T = G_T \bm{\lambda} + g_T\\
    G_N \bm{\lambda} + g_N = 0.
    \end{alignat}
  \end{subequations}
  By differentiating, we get the following system on the derivatives:
  \begin{subequations}
  \begin{alignat}{2}
     K \mathrm{d}\bm{\lambda} =  - \frac{1}{\alpha} H d \bm \sigma_T\\
     d \bm \sigma_T = G_T d \bm \lambda + (dG_T \bm \lambda +dg_T)\\
     G_N\mathrm{d}  \bm{\lambda}   =  - \left(\mathrm{d}G_N \bm{\lambda} + \mathrm{d}g_N \right),
  \end{alignat}
  \end{subequations}
  where $  K= \begin{pmatrix}
         \text{Id}_2 & \mu \bm u_T
     \end{pmatrix} \in \mathbb{R}^{2\times 3}$ and $H = \left (  \text{Id} - \bm u_T  \bm u_T ^\top\right )\in \mathbb{R}^{2\times2}$ with $\bm u_T =  \frac{\bm \sigma_T}{\|\bm \sigma_T\|}$ and $\alpha = \frac{\|\bm \sigma_T \|}{\mu \bm{\lambda}_N}$. We rewrite as 
  \begin{subequations}
  \begin{alignat}{2}
     \left(\frac{1}{\alpha}HG_T + K\right)\mathrm{d}  \bm{\lambda}  = - \frac{1}{\alpha} H \left( \mathrm{d}G_T  \bm{\lambda} + \mathrm{d}g_T\right)\\
     G_N\mathrm{d}  \bm{\lambda}   =  - \left(\mathrm{d}G_N \bm{\lambda} + \mathrm{d}g_N \right).
  \end{alignat}
  \end{subequations}
  Stacking the two equations yields:
  \begin{equation}
      (PG + \Tilde{K})  \mathrm{d}  \bm{\lambda} = -P(\mathrm{d}G  \bm{\lambda} + \mathrm{d}g),\label{eq:syst_in_3}
  \end{equation}
where we introduce $P  =  \begin{pmatrix} \frac{1}{\alpha} H & 0_{2\times 1} \\
  0_{1\times 2} & 1
  \end{pmatrix}
  \in \mathbb{R}^{3\times 3}$ and $\Tilde{K}  =  \begin{pmatrix} K\\
  0_{1\times 3}
  \end{pmatrix}
  \in \mathbb{R}^{3\times 3}$.

Because $\bm \lambda$ is constrained to stay on the boundary of the cone, its variations $\text{d} \bm \lambda$ live in the tangent 2D plane whose basis is $R = \begin{pmatrix}
    \frac{\bm \lambda }{\| \bm \lambda  \|} &  \bm e_z \times \frac{\bm \sigma_T }{\| \bm \sigma_T \|}
\end{pmatrix} \in \mathbb{R}^{3\times2}$.
Applying the change of variable $\mathrm{d}  \bm{\lambda}  =  R\mathrm{d}  \Tilde{\bm{\lambda}}$ and multiplying the previous system \ref{eq:syst_in_3} by $R^\top$ allows getting a linear system of reduced dimension:
\begin{equation}
  \tilde{G} \text{d}\Tilde{\bm{\lambda}} =  - {R}^\top P \left(\mathrm{d}G  \bm{\lambda} + \mathrm{d}g \right),
\end{equation}
where $\tilde{G} = R^\top P G R + Q \in \mathbb{R}^{2 \times 2}$ with $Q =  R^\top \Tilde{K} R =  \begin{pmatrix}
    0 & 0 \\
    0 & 1
\end{pmatrix}\in \mathbb{R}^{2\times 2}$. Hence the final expression when taking derivative w.r.t $\theta$ in the optimal force $\bm \lambda^*$
\begin{equation}
    \tilde{G}  \frac{\mathrm{d}  \bm{\Tilde{\lambda}}}{\text{d} \theta}   =  - {R}^\top P \left(\frac{\mathrm{d}G }{\text{d} \theta} \bm{\lambda}^* + \frac{\mathrm{d}g }{\text{d} \theta}  \right).
\end{equation}

\section{Implicit NCP gradient system}
\label{sec:appendix-NCP_syst}
In this section, we detail the final system solved to compute the gradients of the NCP (equation (10) of the paper).
Following the paper notations, we denote $\mathcal{A}_\text{brk}$, $\mathcal{A}_\text{stk}$, and $\mathcal{A}_\text{sld}$ as the sets of contact indices corresponding to the breaking, sticking, and sliding contacts respectively and $n_\text{brk}$, $n_\text{stk}$, and $n_\text{sld}$ their cardinals and $n = n_\text{brk} + n_\text{stk} + n_\text{sld}$ the total number of contacts.
For the implicit gradients system, we removed the $\text{d} \bm \lambda$ variables associated with contacts in $\mathcal{A}_{brk}$ and sorted the remaining ones by putting first the components associated with the contacts in $\mathcal{A}_{stk}$ before the reduced ones  $\text{d} \Tilde{\bm \lambda}$ of $\mathcal{A}_{sld}$. Then the total variation of $\bm \lambda$ is $\text{d} \bm \lambda = C X \in \RR^{3n}$ with
\begin{align}
&X = \begin{pmatrix}
    \text{d} \bm \lambda^{(1)} \\
    \vdots\\
    \text{d} \bm \lambda^{(n_{\text{stk}})} \\
    \text{d} \Tilde{\bm \lambda}^{(1)} \\
    \vdots\\
    \text{d} \Tilde{\bm \lambda}^{(n_{\text{sld}})}
\end{pmatrix} \in \RR^{3n_{stk} + 2n_{sld}}\\
&C = \left(
    \begin{array}{c|c}
        0_{3n_\text{brk},3n_\text{stk}} & 0_{3n_\text{brk},2n_\text{sld}}\\
        \hline
        \text{Id}_{3n_\text{stk},3n_\text{stk}} & 0_{3n_\text{stk},2n_\text{sld}}\\
        \hline
        0_{3n_\text{sld},3n_\text{stk}} &
            \begin{array}{ccc}
                R^{(1)} & &\\
                & \ddots &\\
                && R^{(n_\text{sld})}
            \end{array}
    \end{array}
\right) \in \RR^{3n\times(3n_{stk} + 2n_{sld})}
\end{align}
Note that in the sticking case, the right-hand side is $\text{d}G \bm \lambda + \text{d}g$ and the left-hand side is $G\text{d}\lambda$ and for the sliding mode, the right-hand side is multiplied by $R^\top P$ and the right-hand side is composed by $R^\top P$ and $R$ plus an additional term due to $Q$. So if we introduce $B$ and $A$ as
\begin{align}
B &= \left(
    \begin{array}{c|c|c}
        0_{3n_\text{stk},3n_\text{brk}} & \text{Id}_{3n_{stk},3n_{stk}} & 0_{3n_\text{stk},3n_\text{sld}}\\
        \hline
        0_{2n_\text{sld},3n_\text{brk}} & 0_{2n_\text{sld},3n_{stk}} &
            \begin{array}{ccc}
                R^{(1)\top}P^{(1)} & &\\
                & \ddots &\\
                && R^{(n_\text{sld})\top}P^{(n_\text{sld})}
            \end{array}
    \end{array}
\right) \in \RR^{ (3n_{stk} + 2n_{sld})\times 3n}\\
A &= BGC + \left(
    \begin{array}{c|c}
        0_{3n_\text{stk},3n_\text{stk}} & 0_{3n_\text{brk},2n_\text{sld}}\\
        \hline
         0_{2n_\text{sld}, 3n_\text{brk}} & 
            \begin{array}{ccc}
                Q & &\\
                & \ddots &\\
                && Q
            \end{array}
    \end{array}
\right) \in \RR^{(3n_{stk} + 2n_{sld})\times  (3n_{stk} + 2n_{sld})}
\end{align}
where $G$ is the complete Delassus matrix that induces coupling between the different contacts.
We recover the linear system of the implicit gradient
\begin{align}
AX = -B(\text{d}G \lambda + \text{d}g),
\end{align}
and, finally, the derivative w.r.t $\theta$ of the force taken in the optimal force $\lambda^*$ as
\begin{align}
\frac{\mathrm{d}  \bm{\lambda}^*}{\text{d} \theta} = - CA^{-1}B\left(\frac{\text{d} G}{\text{d} \theta} \bm{\lambda}^* + \frac{\mathrm{d}g}{\text{d} \theta}\right).
\end{align}

\noindent \textbf{Details on implementation.} In practice, the matrix $B$ and $C$ are not computed, but we directly work on $G$ and $(\text{d}G \lambda + \text{d}g)$ by discarding the right lines and modifying groups of columns and lines to exploit the sparsity of $B$ and $C$. We compute the matrix $A$ and its inverse using a QR decomposition. 

\section{Collision detection contribution}
\label{sec:appendix-coldetection_contrib}
Here, we present the terms $\frac{\partial J_c \bm v^+}{\partial c}\frac{\text{d}c}{\text{d}\theta}$ and 
$\frac{\partial J_c^\top \bm \lambda^*}{\partial c}\frac{\text{d}c}{\text{d}\theta}$ from Section~\ref{subsec:colision-correction}.
Given a contact frame $c$ between body $1$ and body $2$, the contact Jacobian is
\begin{align}
    J_c &= E({}^cX_1J_1 - {}^cX_2J_2),\label{eq:contact_jaco}
\end{align}
where $E = \begin{pmatrix}
    \text{Id}_3 & 0_{3,3}\\0_{3,3}&0_{3,3}
\end{pmatrix}$ is an operator that allows the extraction of the linear part, and $J_1$, $J_2$ are the kinematic Jacobian of the bodies $1$ and $2$.
In practice, the contact placement $c$ is calculated relative the world frame using HPP-FCL. The contact placement is given in function of the placement of two bodies placement relative to the world: ${}^0M_1(\bm q)$ and ${}^0M_2(\bm q)$.
We can write:
\begin{align}
    {}^0M_c(\bm q) &= {}^0\text{CD}({}^0M_1(\bm q), {}^0M_2(\bm q)),
    \label{eq:louis}
\end{align}
where ${}^0\text{CD}$ is an acronym for \textit{collision detection}.
We are interested in the derivatives of $J_c \bm v$  when $J_1$ and $J_2$ are considered constant because their derivation is already considered in the other terms. For clarity of the presentation, we present the calculation for ${}^cX_1J_1 v$ as the derivation for $J_c v$ follows naturally.
In term of Lie groups, ${}^cX_1 = \operatorname{Ad}_{{}^0M_c^{-1}(\bm q){}^0M_1(\bm q)}$
where $\operatorname{Ad}$ denotes the adjoint operator on $\mathrm{SE}(3)$ to explicitly show the dependency in the variables.
With the rules of spatial algebra, we have for a vector $x$ of the lie algebra, and placement $M$, $M'$ in $\mathrm{SE}(3)$
\begin{subequations}
\begin{alignat}{2}
    \text{d} (\operatorname{Ad}_M x) &= - \operatorname{ad}_{ \operatorname{Ad}_M x}\operatorname{Ad}_M\text{d} M\\
    \text{d} (M^{-1} M') &= -\operatorname{Ad}_{M'^{-1} M} \text{d} M + \text{d} M',
\end{alignat}
\end{subequations}
with $\operatorname{ad}$ the small adjoint on the Lie algebra. By the chain rule we obtain
\begin{align}
    \text{d} (\operatorname{Ad}_{M^{-1} M'} x) &= \operatorname{ad}_{ \operatorname{Ad}_{M^{-1} M'} x} \text{d} M - \operatorname{Ad}_{M^{-1} M'}\operatorname{ad}_{x} \text{d} M'.
\end{align}
Applied to $M={}^0M_c(\bm q)$ and $M'={}^0M_1(\bm q)$ we obtain
\begin{align}
    \text{d}({}^cX_1(\bm q)J_1v) &= \operatorname{ad}_{{}^cX_1 J_1 \bm v} \text{d}{}^0M_c -{}^cX_1 \operatorname{ad}_{J_1 \bm v}\text{d} {}^0M_1\\
    &=({}^cX_1 J_1 \bm v)\times \text{d}{}^0M_c - {}^cX_1(J_1 \bm v\times\text{d} {}^0M_1).
\end{align}
Where in the second form we use the generalized cross product from the notation of Featherstone \cite{featherstone2014rigid}.
The calculus is similar for $J_2$.
Now differentiating the collision detection, we have
\begin{align}
    \frac{\text{d}{}^0M_c}{\text{d} \theta} &= \frac{\partial{}^0\text{CD}}{\partial {}^0M_1}J_1 \frac{\text{d} \bm q}{\text{d} \theta} + \frac{\partial{}^0\text{CD}}{\partial {}^0M_2}J_2 \frac{\text{d} \bm q}{\text{d} \theta},
\end{align}
which can be computed using the randomized smoothed derivatives presented in \textit{Differentiable
collision detection: a randomized smoothing approach} [16] in the main paper. For self explanation of the paper, we provide the final formula:
\begin{mdframed}
\begin{align}
\frac{\partial J_c \bm v^+}{\partial c}\frac{\text{d}c}{\text{d}\theta}&=E \left[
    \left(
        ({J_c \bm v^+})_\times \frac{\partial{}^0\text{CD}}{\partial {}^0M_1}
        -({}^cX_1{J_1 \bm v^+})_\times
    \right) J_1
    +   \left(
        ({J_c \bm v^+})_\times \frac{\partial{}^0\text{CD}}{\partial {}^0M_2}
        + ({}^cX_2{J_2 \bm v^+})_\times
    \right) J_2
\right]\frac{\text{d} \bm q}{\text{d} \theta}.\nonumber
\end{align}
\end{mdframed}
To compute the $\frac{\partial J_c^\top \bm \lambda^*}{\partial c}\frac{\text{d}c}{\text{d}\theta}$ we use the previous term and the duality stating that for any $\bm \lambda$ and $\bm v$ we have $\langle J_c^\top \bm \lambda, \bm v \rangle = \langle \bm \lambda, J_c\bm v \rangle$. Taking derivatives we have
\begin{align}
\langle \partial (J_c^\top \bm \lambda) \text{d} \bm q, \bm v \rangle &= \langle \lambda, \partial (J_c \bm v) \text{d} \bm q \rangle\nonumber\\
&= \langle \bm \lambda, L \bm v \rangle\nonumber\\
&= \langle L^\top\bm \lambda,  \bm v \rangle,
\end{align}
and because it is true for any $\bm v$ we have $\partial(J_c^\top \bm \lambda) \text{d} \bm q = L^\top\bm\lambda$.
We calculate $L$ using the anti-commutativity of $\operatorname{ad}$:
\begin{align}
\partial (J_c \bm v) \text{d} \bm q&=E \left[
    \left(
        \operatorname{ad}_{J_c \bm v}\frac{\partial{}^0\text{CD}}{\partial {}^0M_1}
        - {}^cX_1\operatorname{ad}_{J_1 \bm v}
    \right) J_1
    +    \left(
        \operatorname{ad}_{J_c \bm v}\frac{\partial{}^0\text{CD}}{\partial {}^0M_2}
        + {}^cX_2\operatorname{ad}_{J_2 \bm v}
    \right) J_2
\right]\text{d} \bm q\nonumber\\
L \bm v &= - E\left[
    \left(
        \operatorname{ad}_{\frac{\partial{}^0\text{CD}}{\partial {}^0M_1}J_1 \text{d} \bm q}J_c
        - {}^cX_1\operatorname{ad}_{J_1 \text{d} \bm q} J_1
    \right)
    +    \left(
        \operatorname{ad}_{\frac{\partial{}^0\text{CD}}{\partial {}^0M_2}\text{d} \bm q}J_c
        + {}^cX_2\operatorname{ad}_{J_2 \text{d} \bm q} J_2
    \right)
\right]\bm v.
\end{align}
Taking the transpose and introducing the operator $P$ is the variable commutation of $\operatorname{ad}^\top$. Precisely for all $x$ in the Lie algebra and $y$ in the dual Lie algebra: $\operatorname{ad}^\top_x y = \operatorname{P}_y x$ we obtain:
\begin{align}
L^\top \bm \lambda &=-\left[
    \left(
        J_c^\top \operatorname{ad}_{\frac{\partial{}^0\text{CD}}{\partial {}^0M_1}J_1 \text{d} \bm q}^\top
        - J_1^\top\operatorname{ad}^\top_{J_1 \text{d} \bm q} {}^cX_1^\top
    \right)
    +    \left(
        J_c^\top \operatorname{ad}_{\frac{\partial{}^0\text{CD}}{\partial {}^0M_2}J_2 \text{d} \bm q}^\top
        + J_2^\top\operatorname{ad}^\top_{J_2 \text{d} \bm q} {}^cX_2^\top
    \right)
\right]E\bm \lambda\nonumber\\\partial(J_c^\top \bm \lambda) \text{d} \bm q
&= -\left[
    \left(
        J_c^\top \operatorname{P}_{E\bm \lambda}\frac{\partial{}^0\text{CD}}{\partial {}^0M_1}J_1
        - J_1^\top \operatorname{P}_{{}^cX_1^\top E \bm \lambda} J_1
    \right)
    +\left(
        J_c^\top \operatorname{P}_{E\bm \lambda}\frac{\partial{}^0\text{CD}}{\partial {}^0M_2}J_2
        + J_2^\top \operatorname{P}_{{}^cX_2^\top E \bm \lambda} J_2
    \right)
\right] \text{d} \bm q.
\end{align}
And finally, we obtain the second term
\begin{mdframed}
\begin{align}
\frac{\partial J_c^\top \bm \lambda^*}{\partial c}\frac{\text{d}c}{\text{d}\theta}&= \left[
\left(
    J_1^\top \operatorname{P}_{{}^cX_1^\top E \bm \lambda^*}
    - J_c^\top \operatorname{P}_{E\bm \lambda^*}\frac{\partial{}^0\text{CD}}{\partial {}^0M_1}
    \right)J_1
    - \left(
        J_2^\top \operatorname{P}_{{}^cX_2^\top E \bm \lambda^*}
        +         J_c^\top \operatorname{P}_{E\bm \lambda^*}\frac{\partial{}^0\text{CD}}{\partial {}^0M_2}
    \right)J_2
\right]\frac{\text{d} \bm q}{\text{d} \theta}.
\nonumber
\end{align}
\end{mdframed}
\textbf{Details on implementation.} Here, the terms are calculated for one contact. For multiple contacts, $\frac{\partial J_c \bm v^+}{\partial c}\frac{\text{d}c}{\text{d}\theta}$ is the concatenation of the terms for individual contacts and $\frac{\partial J_c^\top \bm \lambda^*}{\partial c}\frac{\text{d}c}{\text{d}\theta}$ is the sum of the terms from each contacts.
Note that for elements of dual spatial algebra $y=[f,m]$ we have the closed form $\operatorname{P}_y  = \begin{pmatrix}
    0 & f_\times\\
    f_\times & m_\times
\end{pmatrix}$.
Note also that similarly to the kinematic Jacobians, the two terms can be computed efficiently by exploiting the sparsity induced by the kinematic structure.

\section{Complete simulation gradients expression}
\label{sec:appendix-gradient_expr}
Combining the terms from the collision detection, the term from the velocity forward kinematic, and the main calculation presented in Section~\ref{sec:gradients}, the complete expression of the simulation step derivative is
\begin{mdframed}
\begin{align}
    \label{eq:total_derivation}
    \frac{\text{d} \bm v^+}{\text{d} \theta} &= \left.\frac{\text{d} \bm v^+}{\text{d} \theta}\right|_{\lambda=\lambda^*} - \Delta t M^{-1}J_c^\top CA^{-1}B
    \left(
    J_c\left.\frac{\text{d} \bm v^+}{\text{d} \theta}\right|_{\lambda=\lambda^*} + \left.\frac{\text{d} J_c \bm v^+}{\text{d} \theta}\right|_{v=v^+} + \frac{\partial J_c \bm v^+}{\partial c}\frac{\text{d}c}{\text{d}\theta}
    \right),\nonumber
\end{align}
\end{mdframed}
with
\begin{align}
    \left.\frac{\text{d} \bm v^+}{\text{d} \theta}\right|_{\lambda=\lambda^*} &=         \frac{\text{d} \bm v}{\text{d}\theta}
        + \Delta t\left(
            \frac{\partial \text{UFD}}{\partial \bm q} \frac{\text{d} \bm q}{\text{d}\theta} 
            + \frac{\partial \text{UFD}}{\partial \bm v} \frac{\text{d} \bm v}{\text{d}\theta}
            + \frac{\partial \text{UFD}}{\partial \bm \tau} \frac{\text{d} \bm \tau}{\text{d}\theta}
            + M^{-1}\frac{\partial J_c^\top \bm \lambda^*}{\partial c}\frac{\text{d}c}{\text{d}\theta}
        \right),\\
    \left.\frac{\text{d} J_c \bm v^+}{\text{d} \theta}\right|_{v=v^+} &= \frac{\partial \text{FKV}(\bm q, \bm v^+, c)}{\partial \bm q} \frac{\text{d} \bm q}{\text{d}\theta}, \label{eq:}
\end{align}
where $\text{FKV}(\bm q, \bm v^+, c)$ is the forward kinematic velocity that gives the velocity of the origin of the frame $c$ when the system is in configuration $\bm q$ with joint velocity $\bm v^+$. $A$, $B$, $C$ are as presented in Appendix B and $\frac{\partial J_c \bm v^+}{\partial c}\frac{\text{d}c}{\text{d}\theta}$, $\frac{\partial J_c^\top \bm \lambda^*}{\partial c}\frac{\text{d}c}{\text{d}\theta}$ are as presented in Appendix C.

\noindent
\textbf{Details on implementation.} 
The partial derivatives $\frac{\partial \text{UFD}}{\partial \bm q, \bm v, \bm \tau}$, $\frac{\partial \text{FKV}}{\partial \bm q}$ and the term $M^{-1}(\bm q) J_c^\top(\bm q)$ can be efficiently computed via rigid-body algorithm as implemented in Pinocchio.
The terms $\frac{\partial J_c \bm v^+}{\partial c}\frac{\text{d}c}{\text{d}\theta}$, $\frac{\partial J_c^\top \bm \lambda^*}{\partial c}\frac{\text{d}c}{\text{d}\theta}$, $A$, $B$ and $C$ can be computed efficiently jointly with the ABA derivatives during forward and backward search of the kinematic tree to exploit its sparsity.

\noindent
\textbf{Baumgarte stabilization} is often used in practice to prevent penetration errors from growing.
The correction is integrated by adding terms to the expression of $g$
\begin{equation}
    g = J_c v_f + \frac{\Phi (q)}{\Delta t} -K_p \left[\frac{\Phi (q)}{\Delta t}\right]_- - K_d J_c v
\end{equation}
where $K_p$ and $K_d$ are the gains of the corrector.
In the case of sticking or sliding contacts we have $(G \bm{\lambda} + g)_N = 0 $ and expanding the expression of $g$ yields
\begin{equation}
    (J_c v^+)_N =  K_d (J_c v)_N - (1-K_p) \frac{\Phi (q)}{\Delta t}.
\end{equation}
Therefore, using a Baumgarte correction affects the derivative of the simulation.
In particular, the derivatives of the proportional term involve $ \Phi$ and thus should be handled when computing the derivatives of the collision detection.
Differentiating the derivative term $ K_d J_c v$ is done similarly to the $ J_c v^+$ term i.e. via the Forward Kinematics derivatives.
In more details, the term in parentheses of \eqref{eq:total_derivation} becomes
\begin{equation}
    \frac{\text{d} G}{\text{d} \theta} + \frac{\text{d} g}{ \text{d} \theta} = J_c\left.\frac{\text{d} \bm v^+}{\text{d} \theta}\right|_{\lambda=\lambda^*} + \left.\frac{\text{d} J_c \bm v^+}{\text{d} \theta}\right|_{v=v^+} + \frac{\partial J_c \bm v^+}{\partial c}\frac{\text{d}c}{\text{d}\theta} +  \frac{(1-K_p) }{\Delta t} \frac{\text{d} \Phi (q)}{\text{d} \theta} - K_d J_c\frac{\text{d} \bm v}{\text{d} \theta} - K_d \left.\frac{\text{d} J_c \bm v}{\text{d} \theta}\right|_{v=v}
\end{equation}

\section{Additional experimental support}
\label{sec:appendix-additional-exp}
\begin{figure}[b]
    \centering
    \includegraphics[width=.25\linewidth]{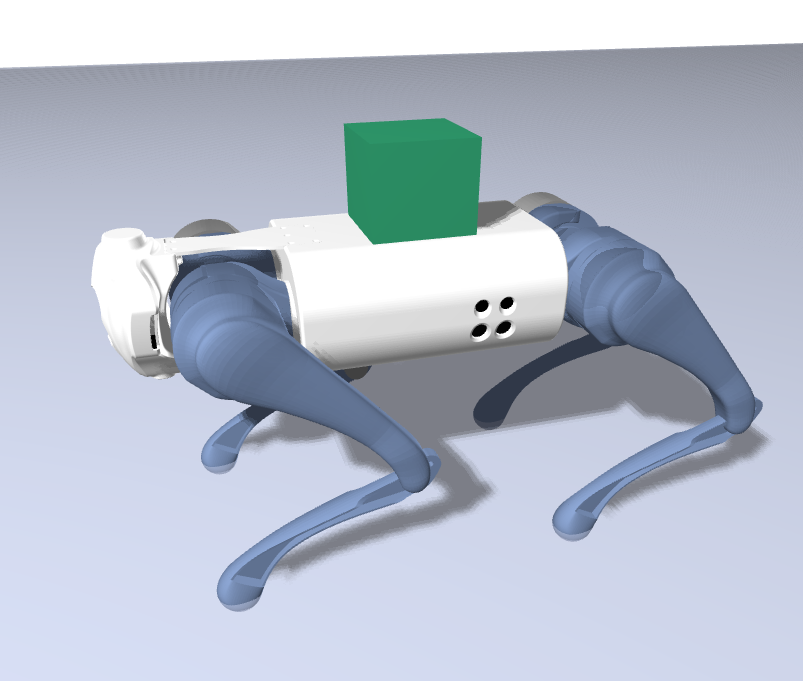}~~
    \includegraphics[width=.25\linewidth]{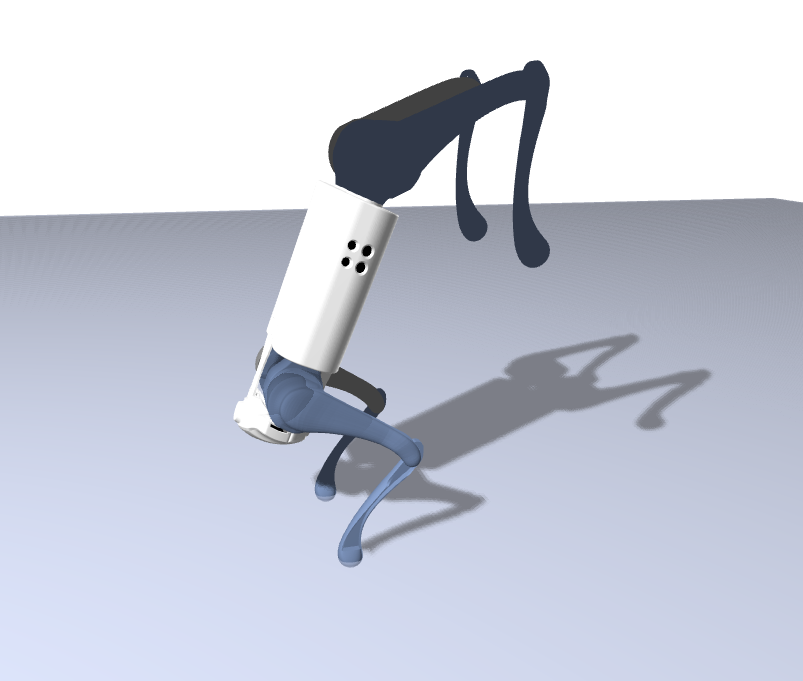}~~
    \includegraphics[width=.25\linewidth]{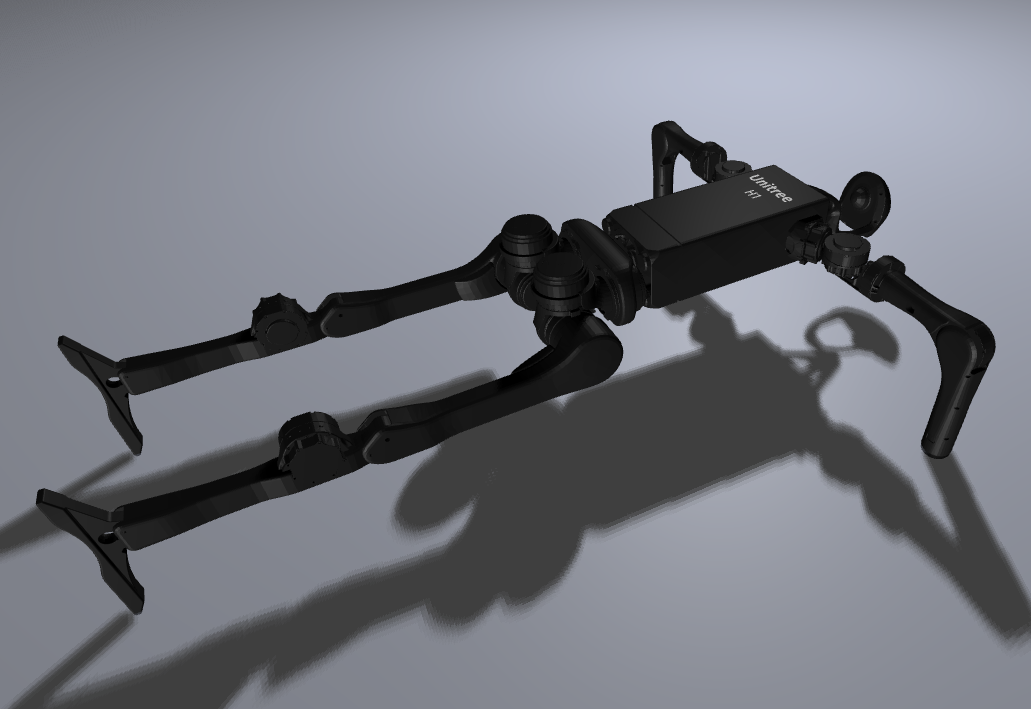}
    \caption{ The differentiable simulator is used to find a control torque stabilizing various robotics systems: a Unitree Go1 in a standing position with a 10kg mass on its back (\textbf{Left}) or in a "hand-stand" pose (\textbf{Center}) and a humanoid Unitree H1 in a "push-up" pose (\textbf{Right}).
    }
    \label{fig:robots_contact_id}
\end{figure}

This section provides several additional experiments using our differentiable simulator to solve contact inverse dynamics problems on underactuated robotics systems.
The considered problems, depicted in Fig.~\ref{fig:robots_contact_id}, are the following:
\begin{itemize}
    \item A Unitree Go1 is stabilized in a standing position with a 10kg mass put on its back;
    \item A Unitree Go1 is stabilized in a "hand-stand" pose;
    \item A Unitree H1 humanoid is stabilized in a "push-up" pose.
\end{itemize}
In every case, the robots are stabilized by optimizing the torque on the actuators.
We refer to the video attached to this paper for more visualization of the experiments.

\section{Policy training details}
\label{sec:appendix-policy}
\noindent
\textbf{CartPole Swing Up} 
task involves stabilizing an underactuated pendulum in an upright position starting from the pendulum hanging downwards. The system comprises a 5-dimensional observation space (cart position $x$ nad velocity $\dot{x}$, pole angle $[\sin(\theta), \cos(\theta)]$ and angular velocity $\dot{\theta}$) and a 1-dimensional action space controlling the cart's prismatic joint torque. The cart joint is constrained to $[-2\text{m}, 2\text{m}]$. and the reward function is defined, similar to \cite{xu2021accelerated}, as:
\begin{equation}
    R = -\theta^2 - 0.1\dot{\theta}^2 - 0.05x^2 - 0.01\dot{x}^2
\end{equation}
Episodes run for 240 time steps without early termination, with randomly sampled initial states.

\noindent
\textbf{Hopper} 
environment evaluates locomotion control of a single-legged robot evolving in a plane. The state space consists of 11 dimensions: base height, rotation, linear velocity (2D) angular velocity; joint angles (3D) and velocities (3D). The action space is 3-dimensional, controlling joint torques for the thigh, leg and foot. The smooth reward function proposed by \cite{xu2021accelerated} combines multiple objectives:
\begin{equation}
    R = R_\text{velocity} + R_\text{height} + R_\text{posture} - 0.1\|a\|^2
\end{equation}
where $R_\text{velocity} = v_x$, and:
\begin{equation}
\begin{aligned}
    R_\text{height} &= \begin{cases}
        -200\Delta_{h}^2, & \text{if } \Delta_h < 0 \\
        \Delta_h, & \text{if } \Delta_h \geq 0
    \end{cases}, \\
    \Delta_h &= \text{clip}(h + 0.3, -1, 0.3)
\end{aligned}
\end{equation}
\begin{equation}
    R_\text{posture} = 1 - \left(\frac{\theta}{30^\circ}\right)^2
\end{equation}
Episodes terminate after 1000 time steps or if the robot height falls below $-0.45\text{m}$.

\noindent
\textbf{PPO hyperparameters}
\begin{table}[h]
\centering
\caption{PPO Hyperparameters}
\label{tab:ppo-params}
\begin{tabular}{r||c|c|c}
\toprule
Environment & Horizon Length & Parallel Envs. & Minibatch Size\\
\midrule
\rowcolor{pastelblue}
CartPole & 240 & 128 & 3840\\
\midrule
Hopper & 32 & 256 & 1024\\
\bottomrule
\end{tabular}
\end{table}
\noindent common to both environments include: $\gamma = 0.99$ and $\lambda = 0.95$ for Generalized Advantage Estimation (GAE) calculation, a learning rate of $3e-4$ for both actor and critic using $10$ mini batch epochs. Environment-specific settings are provided in Table~\ref{tab:ppo-params}.

\noindent
\textbf{SHAC hyperparameters}
\noindent are consistent across both environments: $16$ environments in parallel with a short horizon of $20$, $\gamma = 0.99$ and $\lambda = 0.95$ for GAE calculation, policy and critic learning rates $3e-4$, $16$ training iterations for critic with $4$ minibatches and target value network $\alpha = 99$.

%% file: bibliography.bib
@book{featherstone2014rigid,
  title={Rigid body dynamics algorithms},
  author={Featherstone, Roy},
  year={2014},
  publisher={Springer}
}

@book{griewank2008evaluating,
  title={Evaluating derivatives: principles and techniques of algorithmic differentiation},
  author={Griewank, Andreas and Walther, Andrea},
  year={2008},
  publisher={SIAM}
}

@misc{blondel2024elements,
  title={The elements of differentiable programming},
  author={Blondel, Mathieu and Roulet, Vincent},
  journal={arXiv preprint arXiv:2403.14606},
  year={2024}
}

@inproceedings{mirtich1995impulse,
  title={Impulse-based simulation of rigid bodies},
  author={Mirtich, Brian and Canny, John},
  booktitle={Proceedings of the 1995 symposium on Interactive 3D graphics},
  pages={181--ff},
  year={1995}
}

@inproceedings{elandt2019pressure,
  title={A pressure field model for fast, robust approximation of net contact force and moment between nominally rigid objects},
  author={Elandt, Ryan and Drumwright, Evan and Sherman, Michael and Ruina, Andy},
  booktitle={2019 IEEE/RSJ International Conference on Intelligent Robots and Systems (IROS)},
  pages={8238--8245},
  year={2019},
  organization={IEEE}
}

@book{rawlings2017model,
  author    = {Rawlings, James B. and Mayne, David Q. and Diehl, Moritz M.},
  title     = {Model Predictive Control: Theory, Computation, and Design},
  edition   = {2},
  publisher = {Nob Hill Publishing},
  year      = {2017},
}

@techreport{acary2017contact,
  TITLE = {{On solving contact problems with Coulomb friction: formulations and numerical comparisons}},
  AUTHOR = {Acary, Vincent and Br{\'e}mond, Maurice and Huber, Olivier},
  TYPE = {Research Report},
  NUMBER = {RR-9118},
  PAGES = {224},
  INSTITUTION = {{INRIA}},
  YEAR = {2017},
  MONTH = Nov,
  HAL_ID = {hal-01630836},
  HAL_VERSION = {v1},
}

@article{desaxce1998bipotential,
  TITLE = {{The bipotential method: A constructive approach to design the complete contact law with friction and improved numerical algorithms}},
  AUTHOR = {de Saxc{\'e}, G{\'e}ry and Feng, Z.-Q.},
  JOURNAL = {{Mathematical and Computer Modelling}},
  PUBLISHER = {{Elsevier}},
  VOLUME = {28},
  NUMBER = {4-8},
  PAGES = {225-245},
  YEAR = {1998},
  MONTH = Aug,
  DOI = {10.1016/S0895-7177(98)00119-8},
  HAL_ID = {hal-03883288},
  HAL_VERSION = {v1},
}

@inproceedings{todorov2012mujoco,
  title={Mujoco: A physics engine for model-based control},
  author={Todorov, Emanuel and Erez, Tom and Tassa, Yuval},
  booktitle={2012 IEEE/RSJ international conference on intelligent robots and systems},
  pages={5026--5033},
  year={2012},
  organization={IEEE}
}

@article{lelidec2023contact,
  title={Contact Models in Robotics: A Comparative Analysis},
  author={Quentin Le Lidec and Wilson Jallet and Louis Montaut and Ivan Laptev and Cordelia Schmid and Justin Carpentier},
  journal={IEEE Transactions on Robotics},
  year={2023},
  volume={40},
  pages={3716-3733},
}

@article{simplecontacts2024,
  author    = {Carpentier, Justin and Le Lidec, Quentin and Montaut, Louis},
  title     = {From compliant to rigid contact simulation: a unified and efficient approach.},
  journal   = {Robotics: Science and Systems},
  year      = {2024},
}

@inproceedings{delassus1917memoire,
  title={M{\'e}moire sur la th{\'e}orie des liaisons finies unilat{\'e}rales},
  author={Delassus, {\'E}tienne},
  booktitle={Annales scientifiques de l'{\'E}cole normale sup{\'e}rieure},
  volume={34},
  pages={95--179},
  year={1917}
}

@inproceedings{hu2019difftaichi,
  title={DiffTaichi: Differentiable Programming for Physical Simulation},
  author={Hu, Yuanming and Anderson, Luke and Li, Tzu-Mao and Sun, Qi and Carr, Nathan and Ragan-Kelley, Jonathan and Durand, Fredo},
  booktitle={International Conference on Learning Representations},
  year={2019}
}

@article{geilinger2020add,
  title={Add: Analytically differentiable dynamics for multi-body systems with frictional contact},
  author={Geilinger, Moritz and Hahn, David and Zehnder, Jonas and B{\"a}cher, Moritz and Thomaszewski, Bernhard and Coros, Stelian},
  journal={ACM Transactions on Graphics (TOG)},
  volume={39},
  number={6},
  pages={1--15},
  year={2020},
  publisher={ACM New York, NY, USA}
}

@inproceedings{heiden2021neuralsim,
  author =	  {Heiden, Eric and Millard, David and Coumans, Erwin and Sheng, Yizhou and Sukhatme, Gaurav S},
  year =		  {2021},
  title =		  {Neural{S}im: Augmenting Differentiable Simulators with Neural Networks},
  booktitle = {Proceedings of the IEEE International Conference on Robotics and Automation (ICRA)},
}

@inproceedings{carpentier2018analytical,
  title={Analytical derivatives of rigid body dynamics algorithms},
  author={Carpentier, Justin and Mansard, Nicolas},
  booktitle={Robotics: Science and systems (RSS 2018)},
  year={2018}
}

@inproceedings{montaut2023differentiable,
  title={Differentiable collision detection: a randomized smoothing approach},
  author={Montaut, Louis and Le Lidec, Quentin and Bambade, Antoine and Petrik, Vladimir and Sivic, Josef and Carpentier, Justin},
  booktitle={2023 IEEE International Conference on Robotics and Automation (ICRA)},
  pages={3240--3246},
  year={2023},
  organization={IEEE}
}

@inproceedings{tracy2023differentiable,
  title={Differentiable collision detection for a set of convex primitives},
  author={Tracy, Kevin and Howell, Taylor A and Manchester, Zachary},
  booktitle={2023 IEEE International Conference on Robotics and Automation (ICRA)},
  pages={3663--3670},
  year={2023},
  organization={IEEE}
}

@article{de2018end,
  title={End-to-end differentiable physics for learning and control},
  author={de Avila Belbute-Peres, Filipe and Smith, Kevin and Allen, Kelsey and Tenenbaum, Josh and Kolter, J Zico},
  journal={Advances in neural information processing systems},
  volume={31},
  year={2018}
}

@article{lelidec2021differentiable,
  title={Differentiable simulation for physical system identification},
  author={Le Lidec, Quentin and Kalevatykh, Igor and Laptev, Ivan and Schmid, Cordelia and Carpentier, Justin},
  journal={IEEE Robotics and Automation Letters},
  volume={6},
  number={2},
  pages={3413--3420},
  year={2021},
  publisher={IEEE}
}

@inproceedings{werling2021fast,
  title={Fast and Feature-Complete Differentiable Physics Engine for Articulated Rigid Bodies with Contact Constraints},
  author={Werling, Keenon and Omens, Dalton and Lee, Jeongseok and Exarchos, Ioannis and Liu, C Karen},
  booktitle={Robotics: Science and Systems},
  year={2021}
}

@article{howell2022dojo,
	title={Dojo: {A} {D}ifferentiable {S}imulator for {R}obotics},
	author={Howell, Taylor A. and Le Cleac'h, Simon and Bruedigam, Jan and Kolter, J. Zico and Schwager, Mac and Manchester, Zachary},
	year={2022}
}

@inproceedings{freeman2021brax,
  title={Brax-A Differentiable Physics Engine for Large Scale Rigid Body Simulation},
  author={Freeman, C Daniel and Frey, Erik and Raichuk, Anton and Girgin, Sertan and Mordatch, Igor and Bachem, Olivier},
  booktitle={Thirty-fifth Conference on Neural Information Processing Systems Datasets and Benchmarks Track (Round 1)},
  year={2021}
}

@article{mordatch2012discovery,
  title={Discovery of complex behaviors through contact-invariant optimization},
  author={Mordatch, Igor and Todorov, Emanuel and Popovi{\'c}, Zoran},
  journal={ACM Transactions on Graphics (ToG)},
  volume={31},
  number={4},
  pages={1--8},
  year={2012},
  publisher={ACM New York, NY, USA}
}

@article{kim2023contact,
  title={Contact-implicit mpc: Controlling diverse quadruped motions without pre-planned contact modes or trajectories},
  author={Kim, Gijeong and Kang, Dongyun and Kim, Joon-Ha and Hong, Seungwoo and Park, Hae-Won},
  journal={arXiv preprint arXiv:2312.08961},
  year={2023}
}

@inproceedings{amos2017optnet,
  title={Optnet: Differentiable optimization as a layer in neural networks},
  author={Amos, Brandon and Kolter, J Zico},
  booktitle={International Conference on Machine Learning},
  pages={136--145},
  year={2017},
  organization={PMLR}
}

@article{agrawal2019differentiable,
  title={Differentiable convex optimization layers},
  author={Agrawal, Akshay and Amos, Brandon and Barratt, Shane and Boyd, Stephen and Diamond, Steven and Kolter, J Zico},
  journal={Advances in neural information processing systems},
  volume={32},
  year={2019}
}

@misc{jax2018github,
  author = {James Bradbury and Roy Frostig and Peter Hawkins and Matthew James Johnson and Chris Leary and Dougal Maclaurin and George Necula and Adam Paszke and Jake Vander{P}las and Skye Wanderman-{M}ilne and Qiao Zhang},
  title = {{JAX}: composable transformations of {P}ython+{N}um{P}y programs},
  version = {0.3.13},
  year = {2018},
}

@inproceedings{carpentier2019pinocchio,
   title={The Pinocchio C++ library -- A fast and flexible implementation of rigid body dynamics algorithms and their analytical derivatives},
   author={Carpentier, Justin and Saurel, Guilhem and Buondonno, Gabriele and Mirabel, Joseph and Lamiraux, Florent and Stasse, Olivier and Mansard, Nicolas},
   booktitle={IEEE International Symposium on System Integrations (SII)},
   year={2019}
}

@software{hppfcl,
author = {Pan, Jia and Chitta, Sachin and Manocha, Dinesh and Mirabel, Joseph and Carpentier, Justin and Montaut, Louis},
license = {BSD-2-Clause},
month = jun,
title = {{Coal - An extension of the Flexible Collision Library}},
url = {https://github.com/coal-library/coal},
version = {3.0.4},
year = {2026}
}

@article{gilbert1988fast,
  title={A fast procedure for computing the distance between complex objects in three-dimensional space},
  author={Gilbert, Elmer G and Johnson, Daniel W and Keerthi, S Sathiya},
  journal={IEEE Journal on Robotics and Automation},
  volume={4},
  number={2},
  pages={193--203},
  year={1988},
  publisher={IEEE}
}

@article{van_den_bergen_proximity_2001,
	title = {Proximity {Queries} and {Penetration} {Depth} {Computation} on {3D} {Game} {Objects}},
	journal = {In Game Developers Conference},
	author = {Van den Bergen, Gino},
	year = {2001},
}

@MISC{eigenweb,
  author = {Ga\"{e}l Guennebaud and Beno\^{i}t Jacob and others},
  title = {Eigen v3},
  howpublished = {http://eigen.tuxfamily.org},
  year = {2010}
 }

@article{wei2022coacd,
  title={Approximate convex decomposition for 3d meshes with collision-aware concavity and tree search},
  author={Wei, Xinyue and Liu, Minghua and Ling, Zhan and Su, Hao},
  journal={ACM Transactions on Graphics (TOG)},
  volume={41},
  number={4},
  pages={1--18},
  year={2022},
  publisher={ACM New York, NY, USA}
}

@article{escande2014strictly,
  title={A strictly convex hull for computing proximity distances with continuous gradients},
  author={Escande, Adrien and Miossec, Sylvain and Benallegue, Mehdi and Kheddar, Abderrahmane},
  journal={IEEE Transactions on Robotics},
  volume={30},
  number={3},
  pages={666--678},
  year={2014},
  publisher={IEEE}
}

@article{pang2023global,
  title={Global planning for contact-rich manipulation via local smoothing of quasi-dynamic contact models},
  author={Pang, Tao and Suh, HJ Terry and Yang, Lujie and Tedrake, Russ},
  journal={IEEE Transactions on Robotics},
  year={2023},
  publisher={IEEE}
}

@article{suh2022bundled,
  title={Bundled gradients through contact via randomized smoothing},
  author={Suh, Hyung Ju Terry and Pang, Tao and Tedrake, Russ},
  journal={IEEE Robotics and Automation Letters},
  volume={7},
  number={2},
  pages={4000--4007},
  year={2022},
  publisher={IEEE}
}

@article{lelidec2024leveraging,
  title={Leveraging randomized smoothing for optimal control of nonsmooth dynamical systems},
  author={Le Lidec, Quentin and Schramm, Fabian and Montaut, Louis and Schmid, Cordelia and Laptev, Ivan and Carpentier, Justin},
  journal={Nonlinear Analysis: Hybrid Systems},
  volume={52},
  pages={101468},
  year={2024},
  publisher={Elsevier}
}

@article{tasora2021solving,
  title={Solving variational inequalities and cone complementarity problems in nonsmooth dynamics using the alternating direction method of multipliers},
  author={Tasora, Alessandro and Mangoni, Dario and Benatti, Simone and Garziera, Rinaldo},
  journal={International Journal for Numerical Methods in Engineering},
  volume={122},
  number={16},
  pages={4093--4113},
  year={2021},
  publisher={Wiley Online Library}
}

@article{jourdan1998gauss,
  title={A Gauss-Seidel like algorithm to solve frictional contact problems},
  author={Jourdan, Franck and Alart, Pierre and Jean, Michel},
  journal={Computer methods in applied mechanics and engineering},
  volume={155},
  number={1-2},
  pages={31--47},
  year={1998},
  publisher={Elsevier}
}

@article{della2023model,
  title={{Model-based control of soft robots: A survey of the state of the art and open challenges}},
  author={Della Santina, Cosimo and Duriez, Christian and Rus, Daniela},
  journal={IEEE Control Systems Magazine},
  volume={43},
  number={3},
  pages={30--65},
  year={2023},
  publisher={IEEE}
}

@inproceedings{zhang2023adaptive,
  title={Adaptive barrier smoothing for first-order policy gradient with contact dynamics},
  author={Zhang, Shenao and Jin, Wanxin and Wang, Zhaoran},
  booktitle={International Conference on Machine Learning},
  pages={41219--41243},
  year={2023},
  organization={PMLR}
}

@inproceedings{suh2022differentiable,
  title={Do differentiable simulators give better policy gradients?},
  author={Suh, Hyung Ju and Simchowitz, Max and Zhang, Kaiqing and Tedrake, Russ},
  booktitle={International Conference on Machine Learning},
  pages={20668--20696},
  year={2022},
  organization={PMLR}
}

@article{zhong2022differentiable,
  title={Differentiable Physics Simulations with Contacts: Do They Have Correct Gradients w.r.t. Position, Velocity and Control?},
  author={Zhong, Yaofeng Desmond and Han, Jiequn and Brikis, Georgia Olympia},
  journal={arXiv preprint arXiv:2207.05060},
  year={2022}
}

@ARTICLE{castro2023compliant,
  author={Castro, Alejandro M. and Permenter, Frank N. and Han, Xuchen},
  journal={IEEE Transactions on Robotics}, 
  title={An Unconstrained Convex Formulation of Compliant Contact}, 
  year={2023},
  volume={39},
  number={2},
  pages={1301-1320},
  doi={10.1109/TRO.2022.3209077}}

@INPROCEEDINGS{xu2021diffsim, 
    AUTHOR    = {Jie Xu AND Tao Chen AND Lara Zlokapa AND Michael Foshey AND Wojciech Matusik AND Shinjiro Sueda AND Pulkit Agrawal}, 
    TITLE     = {{An End-to-End Differentiable Framework for Contact-Aware Robot Design}}, 
    BOOKTITLE = {Proceedings of Robotics: Science and Systems}, 
    YEAR      = {2021}, 
    ADDRESS   = {Virtual}, 
    MONTH     = {July}, 
    DOI       = {10.15607/RSS.2021.XVII.008} 
}

@inproceedings{xu2021accelerated,
    title={Accelerated Policy Learning with Parallel Differentiable Simulation},
    author={Xu, Jie and Makoviychuk, Viktor and Narang, Yashraj and Ramos, Fabio and Matusik, Wojciech and Garg, Animesh and Macklin, Miles},
    booktitle={International Conference on Learning Representations},
    year={2022}
  }

@inproceedings{georgiev2024adaptive,
  title={Adaptive Horizon Actor-Critic for Policy Learningin Contact-Rich Differentiable Simulation}, 
  author={Georgiev, Ignat and Srinivasan, Krishnan and Xu, Jie and Heiden, Eric and Garg, Animesh},
  booktitle={International Conference on Machine Learning},
  year={2024},
  organization={PMLR},
}

@inproceedings{Qiao2021Efficient,
author  = {Qiao, Yi-Ling and Liang, Junbang and Koltun, Vladlen and Lin, Ming C.},
title  = {Efficient Differentiable Simulation of Articulated Bodies},
booktitle = {ICML},
year  = {2021},
}

@ARTICLE{hofer2021sim2real,
  author={Höfer, Sebastian and Bekris, Kostas and Handa, Ankur and Gamboa, Juan Camilo and Mozifian, Melissa and Golemo, Florian and Atkeson, Chris and Fox, Dieter and Goldberg, Ken and Leonard, John and Karen Liu, C. and Peters, Jan and Song, Shuran and Welinder, Peter and White, Martha},
  journal={IEEE Transactions on Automation Science and Engineering}, 
  title={Sim2Real in Robotics and Automation: Applications and Challenges}, 
  year={2021},
  volume={18},
  number={2},
  pages={398-400},
  doi={10.1109/TASE.2021.3064065}}

@article{florian2000ipm,
title = {Interior-point methods},
journal = {Journal of Computational and Applied Mathematics},
volume = {124},
number = {1},
pages = {281-302},
year = {2000},
note = {Numerical Analysis 2000. Vol. IV: Optimization and Nonlinear Equations},
issn = {0377-0427},
doi = {https://doi.org/10.1016/S0377-0427(00)00433-7},
author = {Florian A. Potra and Stephen J. Wright},
}

@misc{schulman2017proximalpolicyoptimizationalgorithms,
      title={Proximal Policy Optimization Algorithms}, 
      author={John Schulman and Filip Wolski and Prafulla Dhariwal and Alec Radford and Oleg Klimov},
      year={2017},
      eprint={1707.06347},
      archivePrefix={arXiv},
      primaryClass={cs.LG},
}
